\documentclass[10pt,twocolumn,letterpaper]{article}

\usepackage{cvpr}              

\usepackage{algorithm}
\usepackage{algorithmic}

\usepackage{amsmath}
\usepackage{amssymb}
\usepackage{bm}
\usepackage{tcolorbox}
\usepackage{booktabs}
\usepackage{multirow}

\usepackage{enumitem}
\usepackage{array}
\usepackage{tabularx}
\usepackage{float}
\usepackage[table,xcdraw]{xcolor}

\tcbset{  
  promptbox/.style={  
    colback=white,   
    colframe=white,  
    boxrule=0pt,  
    left=4mm, right=0mm, top=0mm, bottom=0mm,  
    sharp corners  
  }  
}  
\newenvironment{textprompt}{%
  \begin{tcolorbox}[promptbox]  
    \small\ttfamily\raggedright  
}{%
  \end{tcolorbox}  
}

\definecolor{cvprblue}{rgb}{0.21,0.49,0.74}
\usepackage[pagebackref,breaklinks,colorlinks,allcolors=cvprblue]{hyperref}

\def\paperID{N/A} 
\def\confName{CVPR}
\def\confYear{2026}

\title{Hierarchical Textual Knowledge for Enhanced Image Clustering}

\author{Yijie Zhong$^1$, Yunfan Gao$^1$, Weipeng Jiang$^2$, Haofen Wang$^1$\thanks{Corresponding author.}\\
$^1$ Tongji University, $^2$ Huawei Technologies Ltd.\\
{\tt\small dun.haski@gmail.com, 2311821@tongji.edu.cn, jwpqjty@gmail.com, carter.whfcarter@gmail.com}
}

\begin{document}
\maketitle

\begin{abstract}
    Image clustering aims to group images in an unsupervised fashion. Traditional methods focus on knowledge from visual space, making it difficult to distinguish between visually similar but semantically different classes.
    Recent advances in vision-language models enable the use of textual knowledge to enhance image clustering. However, most existing methods rely on coarse class labels or simple nouns, overlooking the rich conceptual and attribute-level semantics embedded in textual space. 
    In this paper, we propose a knowledge-enhanced clustering (KEC) method that constructs a hierarchical concept-attribute structured knowledge with the help of large language models (LLMs) to guide clustering.
    Specifically, we first condense redundant textual labels into abstract concepts and then automatically extract discriminative attributes for each single concept and similar concept pairs, via structured prompts to LLMs. This knowledge is instantiated for each input image to achieve the knowledge-enhanced features. The knowledge-enhanced features with original visual features are adapted to various downstream clustering algorithms.
    We evaluate KEC on 20 diverse datasets, showing consistent improvements across existing methods using additional textual knowledge. KEC without training outperforms zero-shot CLIP on 14 out of 20 datasets. Furthermore, the naive use of textual knowledge may harm clustering performance, while KEC provides both accuracy and robustness.
\end{abstract}

\section{Introduction}

Image clustering aims to group unlabeled images into semantically meaningful clusters~\cite{DBLP:journals/tnn/RenPYXLPYH25}. As a fundamental task in unsupervised learning, it enables the effective organization and understanding of large-scale visual data~\cite{1292402,DBLP:conf/iclr/KwonPKCR024}. The essence of clustering lies in leveraging prior knowledge to establish well-defined class boundaries, thereby achieving clear semantic separation.

Earlier methods primarily rely on knowledge of geometric priors such as connectivity~\cite{DBLP:journals/pami/WangLWNL21} and sparsity~\cite{DBLP:conf/nips/LiuST17} to define structural relationships between samples. These assumptions, though simple and interpretable, struggle with complex visual semantics. 
With the advent of deep learning, clustering has shifted toward leveraging high-dimensional representations learned by CNNs and vision Transformers. These visual patterns serve as implicit knowledge that enables more discriminative clustering~\cite{DBLP:conf/cvpr/ZhangYM24,DBLP:conf/iclr/ChuTDDHV024,DBLP:conf/icml/0008H0WLY24}. Additionally, various pretrained models have been explored to introduce feature-level diversity into the pipeline~\cite{DBLP:conf/icml/GadetskyJB24}.

However, relying solely on visual features often fails to differentiate visually similar but semantically distinct classes~\cite{DBLP:conf/aaai/CaiQ0ZC23}. Recent advances in visual-language models such as CLIP~\cite{DBLP:conf/icml/RadfordKHRGASAM21} have aligned visual and textual spaces, making it possible to inject textual knowledge into clustering. Several recent studies attempt to leverage textual information to enhance clustering performance, but they still face fundamental limitations.

Some approaches directly generate captions or text embeddings for each image using large vision-language models (VLMs)~\cite{DBLP:conf/eacl/StephanMSWGPR24,DBLP:conf/iclr/KwonPKCR024,DBLP:journals/corr/abs-2602-20664}.
While conceptually appealing, this strategy requires repeatedly querying generative models, leading to prohibitive computational overhead and poor scalability. 
Other methods, such as SIC~\cite{DBLP:conf/aaai/CaiQ0ZC23} and TAC~\cite{DBLP:conf/icml/00030P00024}, sidestep this issue by assigning pseudo-labels to images from predefined nouns or lexical resources like WordNet~\cite{DBLP:journals/cacm/Miller95}. Although efficient, these methods rely on shallow textual knowledge, failing to capture deeper semantic hierarchies or discriminative attributes.

\begin{figure*}[t]
    \centering
    \includegraphics[width=\linewidth]{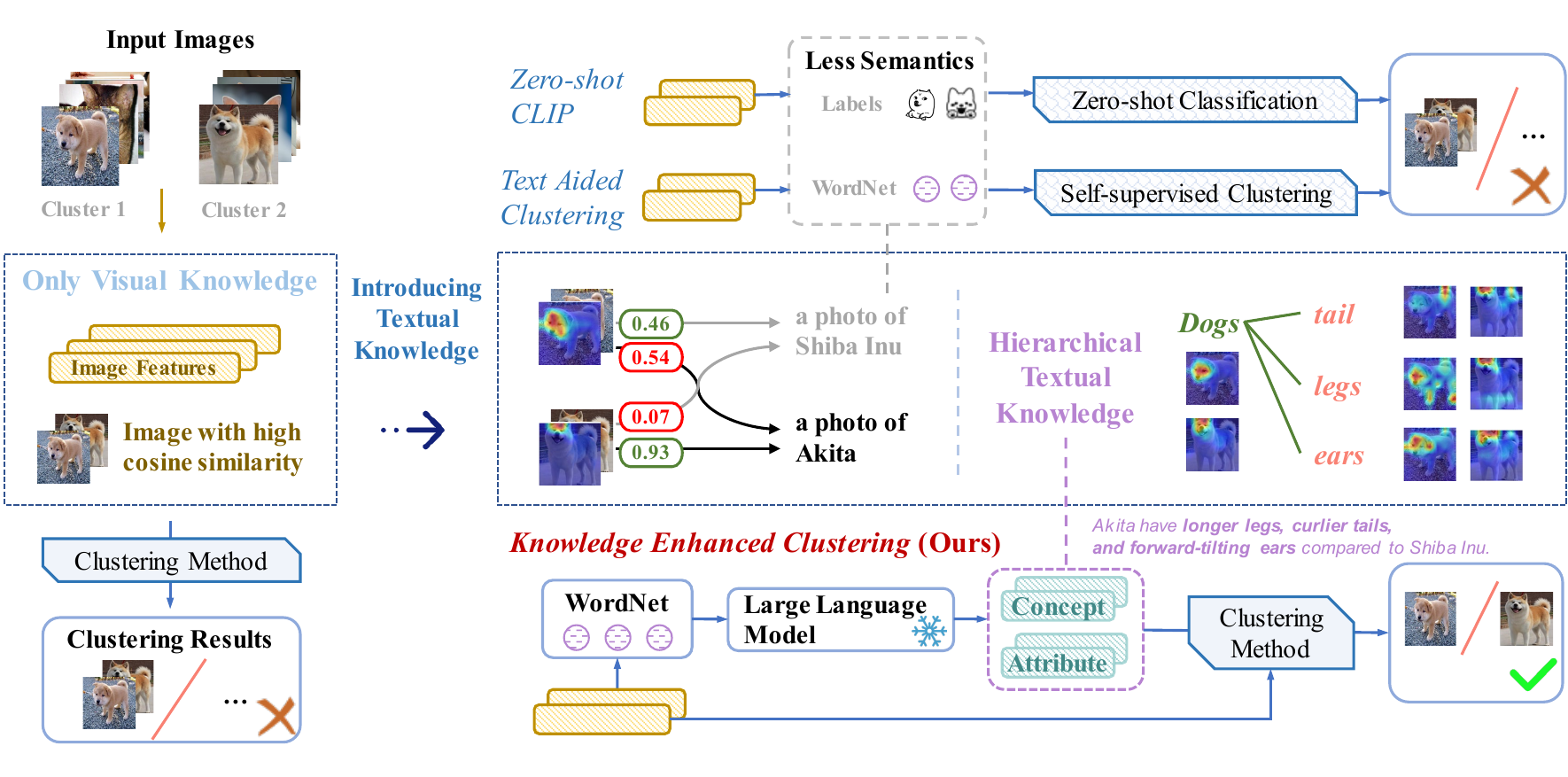}
    \caption{
    Motivation and intuition behind KEC. 
    We first contrast the type of knowledge used by existing methods versus our method. It also illustrates that class labels alone often fail to identify objects of similar categories.
    Building on prior work showing that CLIP attends to targeted semantical regions~\cite{DBLP:conf/cvpr/0002FWZZKXL024}, we visualize the attention maps derived from nouns and discriminative attributes using the method from~\cite{Chefer_2021_ICCV}. The results motivate us to mine discriminative attributes to construct structured textual knowledge for guiding image clustering.}
    \label{fig:intro}
\end{figure*}

In practice, textual knowledge derived in such ways often suffers from semantic redundancy and inconsistent granularity. Similar or synonymous nouns dilute distinctions between similar related categories, while overly broad or narrow terms break the structure of the knowledge base, reducing stability. Consequently, such knowledge in the textual space provides limited guidance for forming clear clustering boundaries. 
In some scenarios, naively incorporating text can even negatively impact clustering performance, underscoring that advances should center on robust textual space modeling rather than on refining noun-selection strategies in previous work.

Drawing inspiration from human cognition, we argue that object differentiation relies not only on category names but also on discriminative attributes. For instance, humans distinguish between an Akita and a Shiba Inu by perceiving subtle differences in leg length, tail curvature, and ear posture. 
As illustrated in~\autoref{fig:intro}, attribute-based cues convey richer semantic information than class names alone, often reflecting accumulated common sense or domain-specific knowledge.
This observation motivates our central question: How can we construct structured and discriminative textual knowledge to guide image clustering? We identify the following two key challenges:

\noindent \textbf{1. How to construct discriminative knowledge in a semantically redundant textual space?}

\noindent \textbf{2. How to acquire such knowledge automatically without relying on manual annotation or feeding raw images into large generative models?}

In this paper, we propose a method called Knowledge-Enhanced Clustering (KEC). KEC leverages large language models to build a hierarchical concept-attribute knowledge structure.
We first abstract representative concepts from redundant noun sets, and then extract discriminative attributes both within and across these concepts. This knowledge is instantiated and grounded in the input images to generate knowledge-enhanced features, which are integrated with visual features for clustering. KEC is fully automated and compatible with various downstream clustering algorithms.
Extensive experiments demonstrate that KEC consistently improves clustering performance and robustness. Moreover, results reveal that naively constructed textual knowledge can degrade performance, while our hierarchical knowledge provides stable and interpretable gains.
The contributions of this work are summarized as follows:
\begin{itemize}
    \item We propose a knowledge-enhanced clustering method that uses large language models to construct structured, hierarchical textual knowledge for guiding clustering.
    \item We introduce a concept-attribute knowledge in textual space. Representative concepts are distilled from redundant nouns, and discriminative attributes are extracted both within and between concepts to explicitly capture differences between similar categories.
    \item We validate our method on 20 visual datasets across diverse scenarios, achieving consistent improvements and robustness. Without any training, KEC outperforms training-based methods by 3\% in NMI on average and surpasses zero-shot CLIP on 14 datasets.
\end{itemize}

\begin{figure*}[t]
    \centering
    \includegraphics[width=\linewidth]{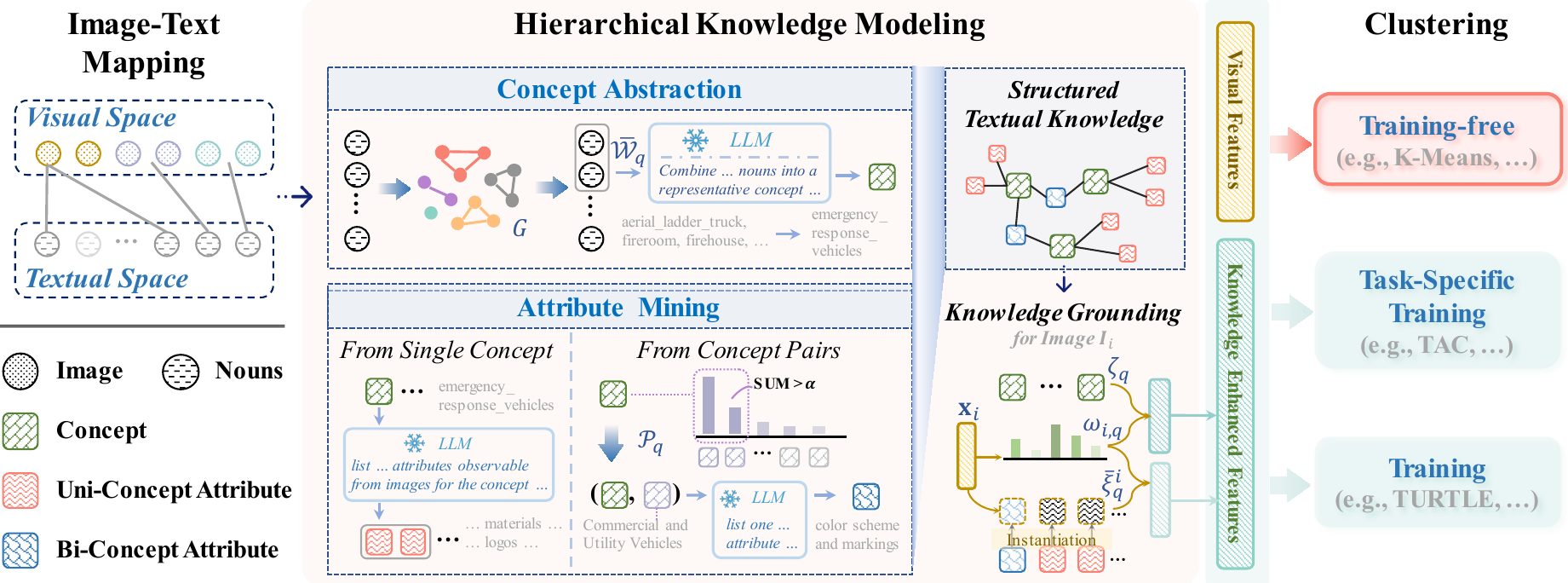}
    \caption{Overview of KEC. 
    Image-Text Mapping first aligns visual features with the textual space using noun features from WordNet. 
    We then construct a hierarchical concept-attribute structured knowledge in the textual space. Representative concepts are distilled from semantically redundant nouns and abstracted using an LLM. Discriminative attributes are generated based on either single concepts or similar concept pairs. After constructing this structured knowledge, KEC instantiates features for each input image to obtain knowledge-enhanced features. 
    These are combined with visual features and can be flexibly integrated into various downstream clustering methods.
    }
    \label{fig:framework}
\end{figure*}

\section{Related Work}

Early clustering methods rely on handcrafted features and geometric assumptions~\cite{DBLP:conf/cvpr/LochmanOZ24}, such as linear subspaces, compactness, connectivity~\cite{DBLP:journals/pami/WangLWNL21}, and sparsity~\cite{DBLP:conf/nips/LiuST17}. These methods define cluster boundaries based on structural priors as knowledge, yet they struggle to handle high-dimensional and semantically rich visual data~\cite{DBLP:conf/iccv/DingTCD0H23}. Subsequent advances incorporate multi-view fusion~\cite{DBLP:journals/pami/ChangCNWL24,ma2026phys} and constrained learning~\cite{DBLP:conf/aaai/GuoJL0X24} to enhance robustness, but still rested on fixed geometric assumptions.

The rise of deep learning enabled data-driven representation learning for clustering. Self-supervised contrastive learning~\cite{DBLP:conf/iclr/ChuTDDHV024,DBLP:journals/mta/ZhongSLSW22}, pseudo-label refinement~\cite{DBLP:conf/nips/LiuHZLP24}, and visual pertaining~\cite{DBLP:conf/icml/GadetskyJB24} have led to more discriminative features, which treat the learned image patterns as visual knowledge.
Hierarchical decoupling in architecture leverages base layers for universal visual features and upper layers for structural separability through graph-cut criteria~\cite{DBLP:conf/accv/HeHMQXL24} or manifold alignment, and differentiable spectral clustering~\cite{DBLP:conf/cvpr/ZhangYM24} or probabilistic merging~\cite{DBLP:conf/cvpr/MetaxasTP23}. This layered approach represents a shift towards explainable systems by explicitly incorporating geometric priors, moving away from black-box models. Despite progress, these methods operate purely in the visual space and are prone to failing in distinguishing visually similar yet semantically different classes.

The emergence of large language models and visual-language models has opened new opportunities. Recent work explores textual knowledge as a complementary modality. Vision-language models like CLIP~\cite{DBLP:conf/icml/RadfordKHRGASAM21} bridge the visual and textual spaces, enabling new clustering paradigms~\cite{DBLP:conf/cvpr/Yao0H24}. 
Some methods rely on image-caption generation~\cite{DBLP:conf/eacl/StephanMSWGPR24} or large vision-language model-driven prompt-based clustering~\cite{DBLP:conf/iclr/KwonPKCR024,DBLP:conf/aaai/GengZYP000W25}. 
However, these methods require repeatedly encoding each image with a generative model, making them computationally expensive and difficult to scale. Other methods like SIC~\cite{DBLP:conf/aaai/CaiQ0ZC23} and TAC~\cite{DBLP:conf/icml/00030P00024} adopt predefined textual resources like WordNet to generate pseudo-labels or filter discriminative nouns. While efficient, these methods rely on static nouns that are shallow, semantically redundant, and lack contextual attributes. 
It is worth noting that while TAC suggests leveraging external knowledge, such as the fact that \textit{`Corgis have shorter and thicker legs compared to Shiba Inu dogs'}, relying solely on filtered nouns does not effectively model this knowledge. In fact, they fail to truly utilize such attribute-level cues.

In contrast to prior work that emphasizes noun selection, KEC shifts the focus toward explicit modeling of the textual space. We establish hierarchical and structured textual knowledge with the help of large language models.
Rather than relying on flat noun sets or user prompts, we automatically extract representative concepts and generate discriminative attributes, capturing both semantic structure and inter-class contrast. Our method is fully automated, scalable, and provides interpretable knowledge that improves clustering performance and robustness.

\section{Method}

Let $\mathcal{I}=\{I_1,\cdots,I_{N_v}\}$ denote the set of $N_v$ input images and let $\mathcal{W}=\{w_1,\cdots,w_{N_t}\}$ represent the set of $N_t$ nouns (\textit{e.g.}, from WordNet~\cite{DBLP:journals/cacm/Miller95}). 
Each image $I_i$ is encoded into a visual embedding $I_i$ using the CLIP vision encoder, and each noun $w_j$ is mapped into a textual embedding $\mathbf{t}_j$ via the CLIP text encoder. This results in two feature sets: visual features $X=\{\mathbf{x}_i\}_{i=1}^{N_v}$ and textual features $T=\{\mathbf{t}_i\}_{i=1}^{N_t}$.
KEC starts with establishing an initial image-text mapping by measuring the similarity between $X$ and $T$, allowing us to identify a subset of nouns relevant to the input images. However, these nouns are still semantically redundant in textual space. Then, KEC leverages large language models to construct a hierarchical knowledge in textual space, as shown in~\autoref{fig:framework}. Specifically, distill semantically redundant nouns into a set of representative concepts, and extract discriminative attributes both from individual concepts and from concept pairs. This knowledge helps clarify differences between similar categories and sharpens cluster boundaries. Next, each image is matched to the concepts, and the associated attributes are instantiated to reflect its visual semantics. The concept features and attribute features together form the knowledge-enhanced textual features. Finally, by integrating the original visual features with the knowledge-enhanced features, KEC supports a wide range of downstream image clustering algorithms.

\subsection{Initializing Image-Text Mapping}

Previous studies~\cite{DBLP:conf/aaai/CaiQ0ZC23,DBLP:conf/icml/00030P00024} have explored various strategies for associating input images with relevant textual semantics. Following the approach of~\cite{DBLP:conf/icml/00030P00024}, we first perform k-means clustering on the visual features $X$, where the number of clusters $k$ is set to $N_v/300$. 
Let $\bm{\mu}_p$ denote the centroid of the $p$-th cluster. For each centroid $\bm{\mu}_p$, we compute the similarity with all noun features $\mathbf{t}_i\in T$, and select the top 5 highest-scoring nouns. Formally, the selected indices $\mathcal{S}_p$ are given by:
\begin{equation}
    \mathcal{S}_p = \mathbf{TopK}(\left\{\bm{\mu}_p^\top\mathbf{t_i}\right\}_{i=1}^{N_t}, 5),
\end{equation}
where $\mathbf{TopK}(\cdot,5)$ returns the indices of the top-5 entries with the highest scores.
The resulting set of nouns in cluster $p$ is denoted as $\mathcal{W}_p=\{w_j|j\in \mathcal{S}_p\}$, and the corresponding textual features as $T_p=\{\mathbf{t}_j|j\in \mathcal{S}_p\}$. These nouns contain redundant semantics, along with overly fine or broad nouns.

\subsection{Representative Concept Abstraction}

In the previous stage, a significantly larger number of clusters than the actual number of semantic classes is employed to preserve fine-grained differences in visual space~\cite{DBLP:conf/icml/00030P00024}. While this `over-classification' strategy improves recall, it also introduces semantic redundancy.
Specifically, the noun list $\mathcal{W}_p$ for cluster $p$ may contain multiple similar or even identical nouns. The semantics within noun lists across clusters are often merely fine-grained variants of the same underlying concept, making it hard to distinguish between synonymous classes.
Moreover, many broader generic nouns do not share the same granularity as these fine-grained nouns.
As a result, incorporating such textual knowledge into clustering may lead to misaligned clustering boundaries and degrade clustering performance.

To address this problem, we merge similar clusters to reduce redundancy. Leveraging LLM's advanced capabilities in fine-grained discrimination, inductive reasoning, and its deep conceptual knowledge, we represent each merged cluster using a representative concept with an associated textual description. This not only enhances the usability but also the interpretability of the merged clusters.

For each cluster, we compute the visual centroid $\mu_p$ and textual centroid $\nu_p$ from $T_p$. We then calculate the visual similarity matrix $R^{vis}_{i,j}=\mathrm{sim}(\mu_i,\mu_j)$ and the textual similarity matrix $R^{text}_{i,j}=\mathrm{sim}(\nu_i,\nu_j)$ for all cluster pairs. The two matrices are integrated through a weighted sum with a parameter $\alpha$ to produce the multi-modal similarity matrix $R_{i,j}$ between clusters:
\begin{equation}
    R_{i,j} = \alpha R^{vis}_{i,j} + (1 - \alpha) R^{text}_{i,j},\qquad i,j = 1,\cdots,k.
\end{equation}
Using a threshold $\beta$, we build an adjacency matrix $G_{i,j}=\bm{1}[R_{i,j}>\beta]$. This matrix is then used to extract connected components $\mathcal{Q}_1,\cdots,\mathcal{Q}_M$ using algorithms such as depth-first search. For each connected component $q$, we aggregate the noun sets, resulting in the merged noun sets as $\overline{\mathcal{W}}_q=\cup_{p\in\mathcal{Q}_q}\mathcal{W}_p$. The collection of these merged sets, $\{\overline{\mathcal{W}_q}\}_{q=1}^M$, is more compact and semantically coherent. And $M$ represents the number of the merged noun sets.

Next, we employ the LLM to generate a representative concept that encompasses all nouns in each merged set $\overline{\mathcal{W}}_q$, along with a concise description that outlines the properties of objects within the concept. The LLM is prompted as:

    
    

\begin{textprompt}
    [Task] Combine the following nouns into a representative **concept** and describe this concept in a few words.
    
    [Additional criteria] It should encompass the given nouns without including other related nouns that do not belong to the same class. ...
    
    [Input] Given Nouns: ...
\end{textprompt}
    
\noindent The outputs provide a concept set $\mathcal{C}=\{c_q\}_{q=1}^M$ and its corresponding description set $\mathcal{D}=\{d_q\}_{q=1}^M$. 
These concepts and descriptions are then fed into the CLIP text encoder, yielding concept features $\phi_q$ and description features $\psi_q$.

\subsection{Discriminative Attribute Mining}

Highly abstract concepts often result in ambiguous clustering boundaries. Humans, however, distinguish between visually similar categories based on discriminative attributes. For instance, Corgis are recognized by their shorter and thicker legs than Shiba Inus. Existing methods that rely solely on class names or nouns fail to capture such cues, as they lack explicit modeling of visual attributes.

To address this, we introduce a mid-level semantic based on discriminative attributes, which sharpens inter-cluster distinctions and improves interpretability. We generate two attribute types using LLMs.

\paragraph{Uni-concept Attribute.}

For each concept $c_q$, we prompt the LLM to identify $\lambda_1$ visual attributes that help differentiate instances of $c_q$ from those of other concepts as follows:

    
    

\begin{textprompt}
    [Task] List the two most **representative** and **distinctive** attributes observable from images for the given concept, which can clearly distinguish objects under this concept from other similar classes.
    
    [Additional criteria] You only need to describe the visual effect of the attribute without providing its function.
    
    [Input] Given concept:
\end{textprompt}

\noindent The generated uni-concept attributes is $\mathcal{U}_q = \{u_q^i\}_{i=1}^{\lambda_1}$.

\paragraph{Bi-concept Attribute.}

Mining attributes across all concept pairs is computationally inefficient and often unnecessary, especially for clearly dissimilar concepts. To focus on genuinely ambiguous cases, we first identify semantically similar concept pairs, which are potentially asymmetric.

For each concept $c_q$, we compute its similarity to every other concept $\pi_{q,l}=\phi_q^\top\phi_l \ (l\neq q)$, and apply normalization to get $\hat\pi_{q,l}=\exp(\pi_{q,l})/\sum_{j\neq q}\exp(\pi_{q,j})$. 
We then sort these similarities in descending order to form a ranking  $\bar{l}_{q,1}, \bar{l}_{q,2}, \cdots$ for $l$. The top-$L_q$ most relevant concepts are selected such that their cumulative similarity exceeds the similarity threshold $\beta$:
\begin{equation}
    \mathcal{N}_q =\{\bar{l}_{q,y} | y = 1, \cdots, L_q\}\}, \mathop{\arg\min}\limits_{L_q} \sum^{L_q}_{i=1}\hat\pi_{q,\bar{l}_{q,i}} \geq \beta
\end{equation}
These neighbors $\mathcal{N}_q$ are paired with $c_q$ to form unordered concept pairs for $c_q$:
    $\mathcal{P}_q = \left\{(c_q, c_l) | l\in \mathcal{N}_q \right\}$.    
For each concept pair $(c_q, c_l)$, the LLM is prompted to provide $\lambda_2$ attributes that highlight the distinctive visual differences between the two concepts. An example prompt is:

        

\begin{textprompt}
    [Task] List one attribute that can distinguish two given concepts, where the specific manifestations of the attribute allow for easy identification of objects belonging to each concept.

    [Additional criteria] ...
        
    [Input] Given two concepts:
\end{textprompt}

\noindent The set of bi-concept attribute between $c_q$ and $c_l$ is denoted as $\mathcal{B}_{q,l}=\{b^i_{q,l}\}_{i=1}^{\lambda_2}$. 
Finally, we construct the full attribute set for concept $c_q$ as $\mathcal{A}_q=\mathcal{U}_q \cup \bigcup_{l\in\mathcal{N}_q}\mathcal{B}_{q,l}$.
All attributes in $\mathcal{A}_q$ are encoded using the CLIP text encoder to obtain attribute features $\xi_{q,i}$, where $i=1,\cdots,|\mathcal{A}_q|$.

\subsection{Knowledge Grounding }

After constructing the hierarchical textual knowledge under a 'concept-attribute' schema, we derive knowledge-enhanced features for each input image $I_i$.
First, we combine the name and its description to form a unified representation for each concept: $\zeta_q = \phi_q + \psi_q$.
Next, we compute the similarity between the image feature $\mathbf{x}_i$ and each $\zeta_q$, and normalize the results to obtain attention weights $\omega_{i,q}=\exp(\mathbf{x}_i^\top\zeta_q)/\sum^M_{l=1}\exp(\mathbf{x}_i^\top\zeta_l)$.
For attributes, we observe that raw attributes alone are insufficient for precise discrimination, as they lack contextual grounding. Instead, we instantiate each attribute based on the visual content: $\hat{\xi}_{q,l}^i=\mathbf{x}_i\odot\xi_{q,l}, l=1,\cdots,|\mathcal{A}_q|$. We then average these instantiated attributes within the same concept to obtain the concept-level attribute features $\bar{\xi}_q^i$. Finally, we compute the weighted concept and attribute features for image $I_i$ as:
\begin{equation}
    \mathbf{c}_i=\sum^M_{q=1}\omega_{i,q}\zeta_q, \quad \mathbf{a}_i = \sum^M_{q=1}\omega_{i,q}\bar{\xi}_q^i.
\end{equation}
The knowledge-enhanced feature is defined as $\kappa_i=\mathbf{c}_i+\mathbf{a}_i$.

\subsection{Clustering}

KEC leverages both the original visual features $\mathbf{x}_i$ and knowledge-enhanced features $\kappa_i$. It can be flexibly integrated into various downstream clustering methods.
In training-free settings, we concatenate the two modalities as $[\mathbf{x}_i,\kappa_i]$ and apply conventional clustering algorithms (\textit{e.g.} K-Means). 
In task-specific representation learning methods (\textit{e.g.} TAC~\cite{DBLP:conf/icml/00030P00024}), the visual and knowledge-enhanced features are treated as complementary views and jointly optimized in a self-supervised fashion.
For methods without task-specific representation learning (\textit{e.g.} TURTLE~\cite{DBLP:conf/icml/GadetskyJB24}), $\mathbf{x}_i$ and $\kappa_i$ are incorporated as features from distinct spaces within the pipeline.

\section{Experiments}

\subsection{Datasets and Evaluation Metrics}

Recent advances in pre-training and clustering have made several commonly used datasets less challenging~\cite{DBLP:conf/icml/00030P00024}. Previous studies~\cite{DBLP:conf/aaai/CaiQ0ZC23,DBLP:conf/icml/00030P00024} typically evaluate their methods on datasets with a limited number of classes. Following the setups in~\cite{DBLP:conf/icml/GadetskyJB24,DBLP:conf/icml/RadfordKHRGASAM21}, we adopt a comprehensive benchmark comprising 20 vision datasets, including several fine-grained classification tasks, to highlight the advantages of knowledge-enhanced information in textual space.
This benchmark covers datasets with diverse class scales: 4 datasets have $1\sim10$ classes, 8 have $10\sim100$ classes, and 8 have over 100 classes. It also features multiple domain-specific datasets, such as those from satellite imagery and medical imaging. 
\textcolor{red}{Further details about the datasets are provided in the Supplementary Material~\autoref{sec:dataset}}.

We use three widely used metrics: accuracy (ACC), Normalized Mutual Information (NMI), and Adjusted Rand Index (ARI). Higher scores signify superior results.

\begin{table*}[t]
\centering
\caption{KEC outperforms existing methods across diverse settings. 
The first three lines reflect the advantages of KEC under training-free settings. 
The last line uses ground-truth class labels as a loosely constrained upper bound. The remaining lines highlight the superiority of our hierarchical textual knowledge under different training strategies. \textbf{Bold} and \underline{underlined} numbers denote the best and second-best results. \textit{Average} refers to the average performance across all 20 datasets.}
\resizebox{6.4in}{!}{
\begin{tabular}{@{}llccclccclccclcccl@{}}
\midrule
\textbf{Dataset} & \textbf{} & \multicolumn{3}{c}{\textbf{Average}} & \textbf{} & \multicolumn{3}{c}{\textbf{Flowers}} & \textbf{} & \multicolumn{3}{c}{\textbf{Pets}} & \textbf{} & \multicolumn{3}{c}{\textbf{CLEVR}} & \textbf{} \\ \cmidrule(lr){3-5} \cmidrule(lr){7-9} \cmidrule(lr){11-13} \cmidrule(lr){15-17}
\textbf{Metrics} & \textbf{} & \textbf{NMI} & \textbf{ACC} & \textbf{ARI} & \textbf{} & \textbf{NMI} & \textbf{ACC} & \textbf{ARI} & \textbf{} & \textbf{NMI} & \textbf{ACC} & \textbf{ARI} & \textbf{} & \textbf{NMI} & \textbf{ACC} & \textbf{ARI} & \textbf{} \\ \midrule
CLIP (k-means) &  & {\color[HTML]{7030A0} 55.6} & {\color[HTML]{7030A0} 53.6} & {\color[HTML]{7030A0} 34.8} &  & 86.5 & 71.5 & \textbf{67.7} &  & 65.0 & 51.3 & 40.5 &  & \textbf{18.9} & \textbf{27.6} & 8.1 &  \\
TAC (no train) &  & {\color[HTML]{7030A0} 55.3} & {\color[HTML]{7030A0} 54.2} & {\color[HTML]{7030A0} 35.1} &  & 84.5 & 69.4 & 64.8 &  & 79.2 & 65.8 & 57.8 &  & 11.8 & 23.6 & 5.4 &  \\
$\mathtt{KEC}$ (no train) &  & {\color[HTML]{7030A0} \textbf{58.0}} & {\color[HTML]{7030A0} \textbf{56.6}} & {\color[HTML]{7030A0} \textbf{38.5}} & \textbf{} & \textbf{87.3} & \textbf{72.8} & 67.5 & \textbf{} & \textbf{81.2} & \textbf{67.8} & \textbf{63.3} &  & 16.3 & 26.6 & \textbf{8.9} &  \\ \midrule \midrule
SIC &  & {\color[HTML]{7030A0} 54.9} & {\color[HTML]{7030A0} 53.0} & {\color[HTML]{7030A0} 35.6} &  & 67.7 & 43.1 & 34.0 &  & 67.7 & 51.6 & 42.6 &  & 5.2 & 20.6 & 1.5 &  \\
TAC &  & {\color[HTML]{7030A0} 55.2} & {\color[HTML]{7030A0} 55.4} & {\color[HTML]{7030A0} 37.2} &  & 80.4 & 64.2 & 59.6 &  & \underline{83.6} & \underline{77.9} & \underline{69.1} &  & 8.4 & 22.4 & 3.6 &  \\
$\mathtt{KEC_{TAC}}$ &  & {\color[HTML]{7030A0} \underline{57.9}} & {\color[HTML]{7030A0} \underline{59.0}} & {\color[HTML]{7030A0} \underline{40.3}} &  & 85.4 & 71.6 & 66.8 &  & \textbf{84.8} & \textbf{79.7} & \textbf{71.5} &  & 15.9 & 25.4 & 6.8 &  \\ \midrule
TURTLE (1-space) &  & {\color[HTML]{7030A0} 56.1} & {\color[HTML]{7030A0} 55.5} & {\color[HTML]{7030A0} 37.8} &  & 90.7 & \underline{87.2} & \underline{79.8} &  & 71.5 & 60.9 & 48.9 &  & 16.0 & 24.6 & 7.0 &  \\
$\mathtt{TAC_{TURTLE}}$ &  & {\color[HTML]{7030A0} 56.2} & {\color[HTML]{7030A0} 55.8} & {\color[HTML]{7030A0} 38.1} &  & \underline{90.8} & 86.8 & 79.1 &  & 73.7 & 63.4 & 52.5 &  & \underline{16.6} & \underline{25.8} & \underline{7.5} &  \\
$\mathtt{KEC_{TURTLE}}$ &  & {\color[HTML]{7030A0} \textbf{58.4}} & {\color[HTML]{7030A0} \textbf{59.3}} & {\color[HTML]{7030A0} \textbf{40.9}} &  & \textbf{92.4} & \textbf{88.4} & \textbf{82.7} &  & 81.0 & 72.4 & 63.3 &  & \textbf{18.8} & \textbf{28.6} & \textbf{8.0} &  \\ \midrule
{\color[HTML]{9B9B9B} CLIP (zero-shot)} & {\color[HTML]{9B9B9B} } & {\color[HTML]{9B9B9B} 55.9} & {\color[HTML]{9B9B9B} 56.7} & {\color[HTML]{9B9B9B} 37.5} & {\color[HTML]{9B9B9B} } & {\color[HTML]{9B9B9B} 79.4} & {\color[HTML]{9B9B9B} 67.5} & {\color[HTML]{9B9B9B} 58.7} & {\color[HTML]{9B9B9B} } & {\color[HTML]{9B9B9B} 87.6} & {\color[HTML]{9B9B9B} 84.9} & {\color[HTML]{9B9B9B} 75.4} & {\color[HTML]{9B9B9B} } & {\color[HTML]{9B9B9B} 16.4} & {\color[HTML]{9B9B9B} 4.4} & {\color[HTML]{9B9B9B} 8.0} & {\color[HTML]{9B9B9B} } \\ 
{\color[HTML]{9B9B9B} VLM Caption + Cluster} & {\color[HTML]{9B9B9B} } & {\color[HTML]{9B9B9B} 50.7} & {\color[HTML]{9B9B9B} 51.9} & {\color[HTML]{9B9B9B} 32.7} & {\color[HTML]{9B9B9B} } & {\color[HTML]{9B9B9B} 60.1} & {\color[HTML]{9B9B9B} 41.3} & {\color[HTML]{9B9B9B} 30.5} & {\color[HTML]{9B9B9B} } & {\color[HTML]{9B9B9B} 56.6} & {\color[HTML]{9B9B9B} 47.8} & {\color[HTML]{9B9B9B} 37.4} & {\color[HTML]{9B9B9B} } & {\color[HTML]{9B9B9B} 5.6} & {\color[HTML]{9B9B9B} 10.8} & {\color[HTML]{9B9B9B} 3.0} & {\color[HTML]{9B9B9B} } \\
\midrule
\end{tabular}}
\label{tab:total}
\end{table*}

\begin{figure*}[t]
    \centering
    \includegraphics[width=\linewidth]{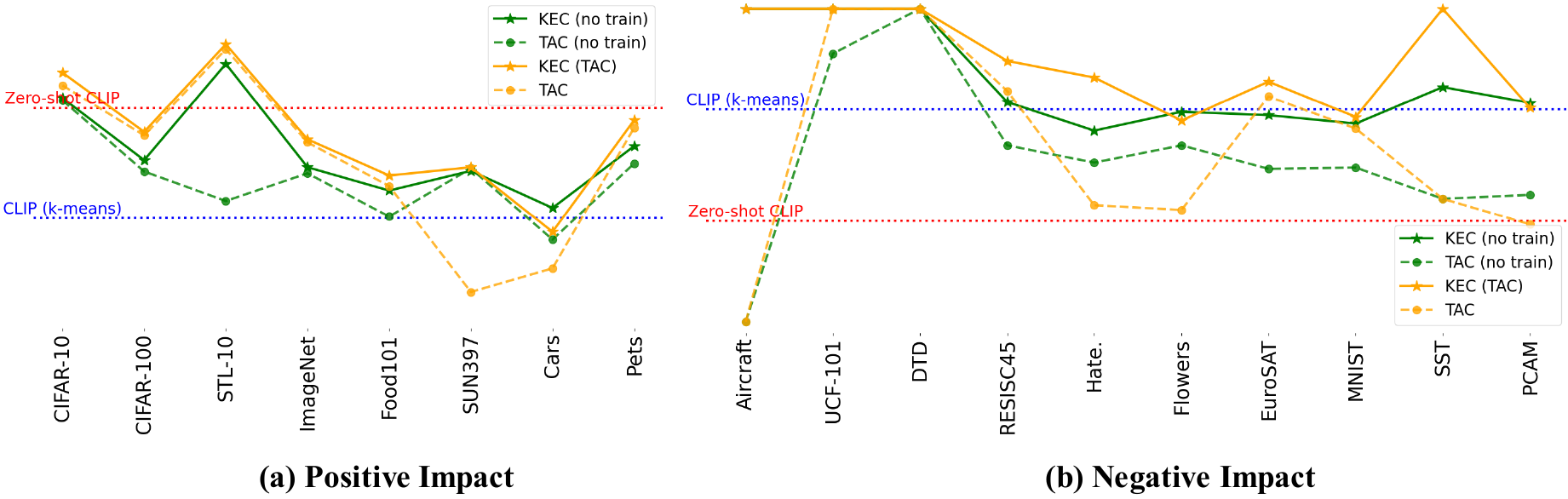}
    \caption{
    Comparisons of different types of textual knowledge. 
    We categorize the impact of textual knowledge into two groups by comparing the performance gap between clustering without textual knowledge and with class label knowledge.
    KEC consistently achieves better or comparable performance by incorporating hierarchical knowledge.
    In contrast, TAC shows limited improvements on a subset of datasets, and its performance trend closely mirrors that of zero-shot CLIP, suggesting that simply using coarse nouns or labels is insufficient and even degrades performance.}
    \label{fig:text}
\end{figure*}

\subsection{Baselines and Implementation Details}

We use CLIP (k-means) and TURTLE (1-space)~\cite{DBLP:conf/icml/GadetskyJB24}, both of which rely solely on visual knowledge, and CLIP (zero-shot), which leverages textual ground-truth class label information, as the baselines. 
Building on advances in incorporating knowledge from the textual space, we also include two state-of-the-art methods: SIC~\cite{DBLP:conf/aaai/CaiQ0ZC23} and TAC (no train)~\cite{DBLP:conf/icml/00030P00024}.
To evaluate the transferability of the knowledge and plug-and-play nature of KEC, we incorporate it into two downstream training strategies: TAC and TURTLE. Specifically, for TAC, we use its original text features or the knowledge-enhanced features from KEC, represented as $\mathtt{TAC}$ or $\mathtt{KEC_{TAC}}$, respectively. For TURTLE, these features are used as additional space features, represented as $\mathtt{TAC_{TURTLE}}$ and $\mathtt{KEC_{TURTLE}}$, respectively. We further include a pure VLM-driven baseline using GPT-4o for image captioning, followed by clustering (VLM Caption + Cluster).

Following previous works~\cite{DBLP:conf/aaai/CaiQ0ZC23,DBLP:conf/icml/00030P00024}, we adopt the pre-trained CLIP ViT-B/32 model to extract visual and textual embeddings. For nouns from WordNet~\cite{DBLP:journals/cacm/Miller95}, we assemble them with prompts like ``A photo of [CLASS]''. We set the parameters $\lambda_1=2,\lambda_2=1$. We set the weight $\alpha$ and threshold $\beta$ to 0.8. All experiments are conducted on an NVIDIA RTX 3090 GPU. We use OpenAI API to access \texttt{GPT-4o} for all LLM needs. 
\textcolor{red}{More details are provided in the Supplementary Material~\autoref{sec:detail}.} 

{\renewcommand{\arraystretch}{1} 
\begin{table*}[t]
\centering
\caption{Ablation studies of different levels of the knowledge and their components in KEC. 
\textit{Con.} and \textit{Des.} mean the name of the concept and its corresponding description. \textit{UA} and \textit{BA} mean the uni-concept attribute and bi-concept attribute, respectively.}
\resizebox{6.4in}{!}{
\begin{tabular}{@{}cccclcccccccccccccccl@{}}
\midrule
 &  &  &  &  & \multicolumn{3}{c}{\textbf{Average}} & \multicolumn{1}{l}{} & \multicolumn{3}{c}{\textbf{Flowers}} & \multicolumn{1}{l}{} & \multicolumn{3}{c}{\textbf{Pets}} & \multicolumn{1}{l}{} & \multicolumn{3}{c}{\textbf{CLEVR}} &  \\ \cmidrule(lr){6-8} \cmidrule(lr){10-12} \cmidrule(lr){14-16} \cmidrule(lr){18-20}
\multirow{-2}{*}{\textbf{Con.}} & \multirow{-2}{*}{\textbf{Des.}} & \multirow{-2}{*}{\textbf{UA}} & \multirow{-2}{*}{\textbf{BA}} &  & \textbf{NMI} & \textbf{ACC} & \textbf{ARI} & \multicolumn{1}{l}{} & \textbf{NMI} & \textbf{ACC} & \textbf{ARI} & \multicolumn{1}{l}{} & \textbf{NMI} & \textbf{ACC} & \textbf{ARI} & \multicolumn{1}{l}{} & \textbf{NMI} & \textbf{ACC} & \textbf{ARI} &  \\ \midrule
\checkmark &  &  &  &  & {\color[HTML]{7030A0} 56.5} & {\color[HTML]{7030A0} 54.9} & {\color[HTML]{7030A0} 36.9} &  & 84.9 & 68.9 & 62.3 &  & 78.9 & 66.0 & 61.1 &  & 14.2 & 23.2 & 5.5 &  \\
 & \checkmark &  &  &  & {\color[HTML]{7030A0} 56.2} & {\color[HTML]{7030A0} 54.8} & {\color[HTML]{7030A0} 36.4} &  & 85.0 & 68.6 & 62.0 &  & 78.9 & 64.9 & 60.9 &  & 15.4 & 24.2 & 6.3 &  \\
\checkmark & \checkmark &  &  &  & {\color[HTML]{7030A0} 56.7} & {\color[HTML]{7030A0} 55.4} & {\color[HTML]{7030A0} 37.2} &  & 85.5 & 70.8 & 63.6 &  & 79.0 & 64.7 & 60.5 &  & 15.8 & 25.8 & 8.0 &  \\ \midrule
\checkmark & \checkmark & \checkmark & \multicolumn{1}{l}{} &  & {\color[HTML]{7030A0} 57.8} & {\color[HTML]{7030A0} 56.3} & {\color[HTML]{7030A0} 38.3} &  & 87.3 & 72.6 & 67.6 &  & 80.5 & 67.3 & 63.1 &  & 16.0 & 26.6 & 8.6 &  \\
\checkmark & \checkmark & \multicolumn{1}{l}{} & \checkmark &  & {\color[HTML]{7030A0} 57.8} & {\color[HTML]{7030A0} 56.2} & {\color[HTML]{7030A0} 38.1} &  & 86.8 & 72.2 & 65.2 &  & 80.2 & 66.7 & 62.8 &  & 16.3 & 26.6 & 8.9 &  \\
\checkmark & \checkmark & \checkmark & \checkmark &  & {\color[HTML]{7030A0} 58.0} & {\color[HTML]{7030A0} 56.6} & {\color[HTML]{7030A0} 38.5} &  & 87.3 & 72.5 & 67.5 &  & 81.2 & 67.8 & 63.3 &  & 16.3 & 26.6 & 8.9 &  \\ \midrule
\end{tabular}}
\label{tab:ab}
\end{table*}}

\begin{figure*}[t]
    \centering
    \includegraphics[width=\linewidth]{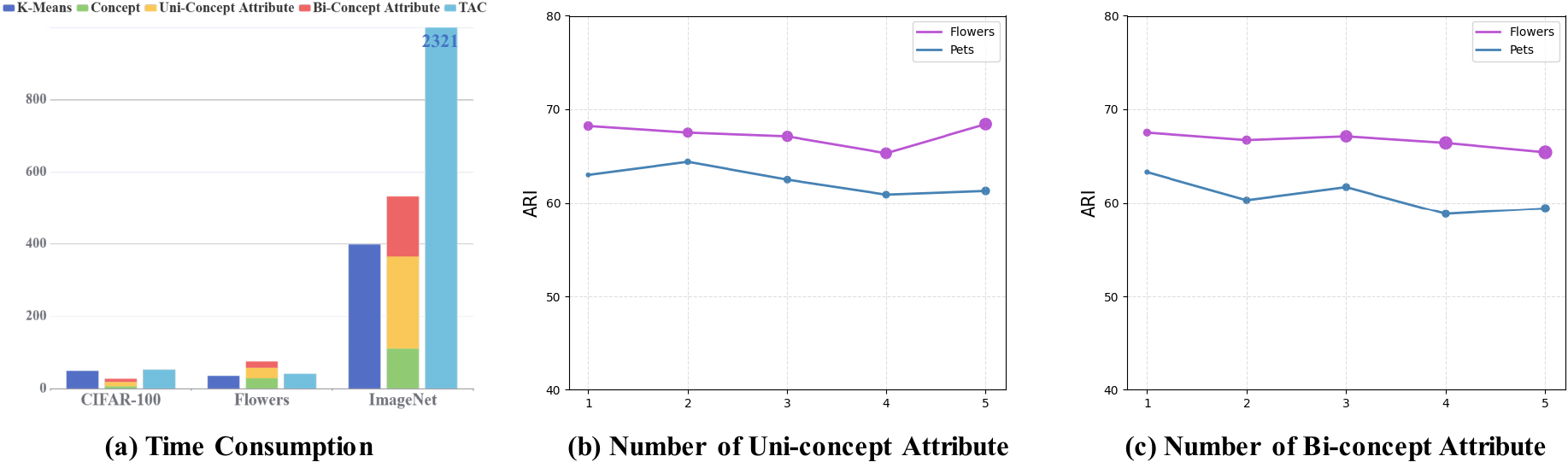}
    \caption{Analyses on time consumption and hyperparameters. 
    (a) Runtime comparison between KEC and TAC in the knowledge construction stage across datasets with varying numbers of classes and images.
    We break down the runtime of KEC into three components (Concept, Uni-Concept Attribute, and Bi-Concept Attribute) and include K-Means as a lightweight downstream clustering algorithm for reference.
    (b) and (c) show the impact of the number of attributes on clustering performance (measured by ARI) across datasets of different scales. The size of each point reflects the corresponding runtime, reflecting the relation between attribute numbers and computational cost.
    }
    \label{fig:param}
\end{figure*}

\subsection{Main Results}

\paragraph{The Superiority of Constructed Knowledge in KEC.}

\autoref{tab:total} summarizes the comparison between KEC and existing methods. 
We report the average performance on the benchmark of 20 datasets. Also, we select three datasets containing different scales of classes: Flowers (102 classes)~\cite{DBLP:conf/icvgip/NilsbackZ08}, Pets (37 classes)~\cite{DBLP:conf/cvpr/ParkhiVZJ12}, and CLEVR (8 classes)~\cite{DBLP:conf/cvpr/JohnsonHMFZG17}. \textcolor{red}{The complete results are provided in 
the Supplementary Material~\autoref{sec:all}.}
We first compare KEC with other approaches for constructing knowledge in textual space (pseudo-labels for SIC and noun combinations for TAC). KEC achieves an improvement of $\sim3\%$ across all metrics in the average performance, with a notable $\sim10\%$ ARI increase on the Pets dataset.
It could be found that KEC outperforms the existing methods under a no-training setting, even surpassing zero-shot CLIP and methods that rely on training with constructed knowledge. 
Further training leveraging KEC's constructed knowledge with the help of LLM substantially enhances clustering performance.
Furthermore, the knowledge in KEC transfers effectively, achieving consistent performance gains across diverse downstream training strategies. This demonstrates the strong usability of our method. The VLM-driven method achieves an overall accuracy 4.7\% lower than KEC and is computationally costly (taking several seconds per image), as per-image VLM calls overlook inter-image contrasts and dwell on irrelevant details. This highlights the importance of explicit textual modeling in KEC.

\paragraph{The Robustness of Constructed Knowledge in KEC.}
Extensive experiments on a broader range of datasets indicate that simply incorporating knowledge in textual space does not always improve clustering performance, as evidenced in~\autoref{tab:total}. This finding motivates a deeper investigation into the robustness of textual knowledge in image clustering. We assess the impact of textual knowledge by comparing performance using textual knowledge against visual knowledge alone, and the results are shown in~\autoref{fig:text}. Then we compared different ways of constructing textual knowledge. TAC shows a performance trend consistent with zero-shot CLIP. Relying solely on a single type of knowledge fails to adapt to various scenarios. Even when the knowledge in TAC has a positive impact, the improvements are limited. And in many cases, it degrades clustering performance. In contrast, KEC builds hierarchical textual knowledge to achieve better results consistently, alleviating concerns about performance loss. This demonstrates the strong robustness and usability of our proposed method.

We further explore enhanced knowledge construction to build a more precise textual space. A human-machine interaction is introduced to guide the model toward attributes and concepts linked to specific categories or high-level semantics (\textit{e.g.} aircraft and hatefulmeme), denoted as KEC$^+$. This approach enables KEC to capture more discriminative semantics, resulting in a 9.4\% accuracy improvement on Hatefulmeme. \textcolor{red}{More results are provided in \autoref{sec:all}}.

\paragraph{The Efficiency of Knowledge Construction in KEC.}
\autoref{fig:param}(a) details the time consumption of each component and the overall process of knowledge construction in KEC. Results are shown for CIFAR-100 (100 classes, 60k images), Flowers (102 classes, 8k images), and ImageNet (1k classes, 1.32M images), representing datasets of varying scales.
Compared to TAC, our method has a similar time consumption on small-scale datasets. However, as the size of the dataset increases, KEC proves to be more efficient.
Since KEC could integrate with various downstream clustering algorithms, we select the simplest K-Means for further runtime comparison. The results show that KEC's computational cost remains close to that of K-Means across different dataset sizes.
This indicates KEC improves performance at an acceptable cost for knowledge construction.

\subsection{Ablation Study}

We conduct experiments by progressively adding different levels of knowledge, including representative concepts and discriminative attributes. Furthermore, we validate the roles of the components at each level. The results are shown in~\autoref{tab:ab}, with \textcolor{red}{complete numerical results available in the Supplementary Material~\autoref{sec:abl}}.

According to Line 1, using only the generated concepts exhibits strong performance. In some cases (\textit{e.g.} CLEVR), the descriptions of the concepts enable KEC to separate similar concepts and improve clustering as shown in Line 2. From Line 3, combining concepts with their descriptions brings broader applicability and further improvements.

Discriminative attributes leverage the knowledge within LLMs, boosting the performance of KEC by over 1\% on all metrics across 20 datasets, with a significant 5\% gain on Flowers.
Comparing Line 4 and Line 5, uni-concept attributes and bi-concept attributes serve different purposes and achieve improvements in varying degrees. Uni-concept attributes capture the signature features of each concept, while bi-concept attributes explore the dimensions to best distinguish similar concepts.
As evidenced in Line 6, uni- and bi-concept attributes complement each other to refine the clustering boundaries and achieve optimal performance.

\subsection{Parameter Analyses}

\paragraph{Number of Uni-concept Attribute $\lambda_1$.} 
To better capture concept semantics and form clear clustering boundaries, KEC identifies $\lambda_1$ discriminative attributes per concept. 
As shown in~\autoref{fig:param}(b), KEC remains robust across different $\lambda_1$ values.
Increasing the number of attributes can initially improve the performance. However, too many attributes degrade clustering quality. Although performance slightly recovers with more attributes (reaching 5), the added computational cost and time outweigh the benefits.
Therefore, in practice, we use $\lambda_1=2$ for a good balance across datasets.

\paragraph{Number of Bi-concept Attribute $\lambda_2$.}
KEC mines the most discriminative attributes for similar concept pairs to distinguish similar concepts and avoid boundary ambiguity.
Here, we evaluate the impact of different numbers $\lambda_2$ of bi-concept attributes, with results shown in~\autoref{fig:param}(c).
The results show that as the number of attributes increases, clustering performance consistently declines across datasets. This aligns with human intuition: distinguishing objects that belong to similar categories often requires focusing on only a limited number of key attributes. Excessive perspectives may introduce noise, leading to erroneous judgments. Therefore, we set the default value of $\lambda_2$ to 1, which is both optimal and the most efficient. \\

\noindent \textcolor{red}{Supplementary Material~\autoref{sec:abl} offers more analyses of parameter selection and pretrained model and LLM choices}. Specifically, we analyze the parameters in Image-Text Mapping, the weight to balance visual and textual similarity, and the threshold in knowledge construction. KEC applies a single configuration and generalizes well, indicating improvement is not due to hyperparameter tuning. We also conduct experiments on various LLMs (from the 0.6B model to the flagship models), and KEC achieves a robustness improvement with the model-agnostic prompts. 

\section{Conclusion}

In this paper, we present a knowledge-enhanced image clustering method (KEC) that leverages LLMs to construct a hierarchical concept-attribute structured knowledge. By distilling redundant WordNet nouns into representative concepts and extracting discriminative attributes, KEC introduces rich semantic knowledge into the textual space without relying on image inputs or generative models. Extensive experiments across 20 datasets validate the effectiveness, robustness, and generalizability of KEC, showing consistent improvements over existing methods.
Our findings also highlight the limitations of shallow textual knowledge and demonstrate the potential of structured textual knowledge as a reliable source for unsupervised visual understanding.

\noindent \textbf{Acknowledgement.}
This work was supported by the National Natural Science Foundation of China (U23B2057).

{
    \small
    \bibliographystyle{ieeenat_fullname}
    \bibliography{main}
}

\clearpage
\setcounter{page}{1}
\maketitlesupplementary

{\renewcommand{\arraystretch}{1.05} 
\begin{table*}[!h]
\centering
\caption{The summary of the used symbols.}
\resizebox{\linewidth}{!}{
\begin{tabular}{p{0.15\linewidth}|p{0.75\linewidth}}
\hline
\multicolumn{1}{l|}{\textbf{Symbol}} & \textbf{Meaning} \\ \hline
\multicolumn{2}{l}{\textit{Initializing Image-Text Mapping}} \\ \hline
\multicolumn{1}{l|}{$\mathcal{I}, I_{i}$} & the collection of input images and a single input image \\
\multicolumn{1}{l|}{$\mathcal{W}, w_{i}$} & the collection of nouns and a single noun \\
\multicolumn{1}{l|}{$N_v,N_t$} & the number of input images and nouns \\
\multicolumn{1}{l|}{$X,\mathbf{x}_i$} & the collection of visual features and the feature of image $I_i$ \\
\multicolumn{1}{l|}{$T,\mathbf{t}_i$} & the collection of noun features and the feature of noun $w_i$ \\ \hline
\multicolumn{2}{l}{\textit{Representative Concept Construction}} \\ \hline
\multicolumn{1}{l|}{$\bm{\mu}_p$} & the centroid of cluster $p$ \\
\multicolumn{1}{l|}{$\mathcal{S}_p$} & Index of the highest scoring noun in cluster $p$ \\
\multicolumn{1}{l|}{$W_p, T_p$} & the collection of nouns and their features of cluster $p$ \\
\multicolumn{1}{l|}{$R_{i,j}$} & the similarity between cluster $i$ and cluster $j$ \\
\multicolumn{1}{l|}{$G, Q$} & the adjacency matrix and the collection of connected components of clusters \\
\multicolumn{1}{l|}{$\mathcal{C}, c_q$} & the collection of concept and a single concept \\
\multicolumn{1}{l|}{$\mathcal{D}, d_q$} & the collection of description of concept and a single description \\
\multicolumn{1}{l|}{$\phi_q, \psi_q$} & the feature of concept and its description \\ \hline
\multicolumn{2}{l}{\textit{Discriminative Attribute Construction}} \\ \hline
\multicolumn{1}{l|}{$\lambda_1,\lambda_2$} & the nubmer of uni-concept attribute and bi-concept attribute \\
\multicolumn{1}{l|}{$\mathcal{U}_q, u^i_q$} & the collection of uni-concept attribute for concept $c_q$ and the i-th attribute \\
\multicolumn{1}{l|}{$\pi_{q,l}$} & the normalized similarity between concept $c_q$ and $c_l$ \\
\multicolumn{1}{l|}{$\bar{l}_{q,1}, \bar{l}_{q,2}, \cdots$} & the index of $l$ for the sorted $\pi_{q,l}$ \\
\multicolumn{1}{l|}{$\mathcal{P}_q$} & the concept pair of concept $c_q$ \\
\multicolumn{1}{l|}{$\mathcal{B}_{q,l}$} & the bi-concept attributes for concept pair ($c_q$, $c_l$) \\
\multicolumn{1}{l|}{$\mathcal{A}_q, \xi_{q,i}$} & the collection of attributes of concept $c_q$ and the i-th attribute \\ \hline
\multicolumn{2}{l}{\textit{Knowledge-enhanced Feature}} \\ \hline
\multicolumn{1}{l|}{$\zeta_q$} & the concept feature for conecpt $c_q$ \\
\multicolumn{1}{l|}{$\bar{\xi}_q^i$} & the average of instantiated attribute features of concept $c_q$ \\
\multicolumn{1}{l|}{$\omega_{i,q}$} & the attention weight between the i-th input image and concept $c_q$ \\
\multicolumn{1}{l|}{$\mathbf{c}_i, \mathbf{a}_i$} & the concept and attribute feature of the i-th input image $I_i$ \\
\multicolumn{1}{l|}{$\kappa_i$} & the knowledge-enhanced feature of the i-th input image $I_i$ \\ \hline
\end{tabular}}
\label{tab:symbol}
\end{table*}}

{\renewcommand{\arraystretch}{1.0} 
\begin{table*}[!h]
\centering
\caption{Details of 20 datasets.}
\resizebox{\linewidth}{!}{
\begin{tabular}{@{}lccc|lccc@{}}
\toprule
\textbf{Dataset} & \textbf{Number fo Classes} & \textbf{Train Size} & \textbf{Test Size} & \textbf{Dataset} & \textbf{Number fo Classes} & \textbf{Train Size} & \textbf{Test Size} \\ \midrule
CIFAR-10 & 10 & 50,000 & 10,000 & DTD & 47 & 3,760 & 1,880 \\
CIFAR-100 & 100 & 50,000 & 10,000 & SUN397 & 397 & 19,850 & 19,850 \\
STL-10 & 10 & 5,000 & 8,000 & EuroSAT & 10 & 10,000 & 5,000 \\
ImageNet & 1000 & 1,281,167 & 50,000 & Resisc45 & 45 & 25,200 & 6,300 \\
Food101 & 101 & 75,750 & 25,250 & GTSRB & 43 & 26,640 & 12,630 \\
Flowers & 102 & 2,040 & 6,149 & PCAM & 2 & 294,192 & 32,768 \\
Cars & 196 & 8,144 & 8,041 & UCF101 & 101 & 9,537 & 3,783 \\
Aircraft & 100 & 6,667 & 3,333 & CLEVR & 8 & 2,000 & 500 \\
Pets & 37 & 3,680 & 3,669 & HatefulMemes & 2 & 8,500 & 500 \\
MNIST & 10 & 60,000 & 10,000 & SST & 2 & 7,792 & 1,821 \\ \bottomrule
\end{tabular}}
\label{tab:dataset}
\end{table*}}

\section{Content List}

We provide additional details and results to complement the main paper. It is organized as follows:

\begin{itemize}
    \item \autoref{sec:symbol} lists all notations and their meanings.
    \item \autoref{sec:dataset} provides the details of 20 datasets used.
    \item \autoref{sec:detail} describes implementation details of all the compared methods and the proposed method KEC.
    \item \autoref{sec:all} presents complete results across all datasets in comparison with existing methods. We also explore further improvement with human interaction, named KEC$^+$.
    \item \autoref{sec:abl} presents a more comprehensive set of ablation studies, including the analyses of different modules, parameter selection, clustering methods, pretrained model choices, and various LLMs. 
    \item \autoref{sec:know} describes the constructed knowledge space.
    \item \autoref{sec:limit} discusses the limitations of our work and outlines future work possibilities.
\end{itemize}

\section{Symbol Definitions}
\label{sec:symbol}

We list the symbols used throughout this paper along with their meanings in~\autoref{tab:symbol}. We hope this could assist readers in better understanding the items presented in our work.

\section{More Details of the Datasets}
\label{sec:dataset}

We evaluate our knowledge-enhanced clustering method on 20 vision datasets. Details of each dataset are provided in~\autoref{tab:dataset}. These datasets cover a wide range of vision tasks, including:
\begin{itemize}[leftmargin=*]
    \item General object classification datasets: CIFAR-10~\cite{2009Learning}, CIFAR-100~\cite{2009Learning}, STL-10~\cite{DBLP:journals/jmlr/CoatesNL11}, ImageNet~\cite{DBLP:conf/cvpr/DengDSLL009};
    \item Fine-grained object classification datasets: \\ Food101~\cite{DBLP:conf/eccv/BossardGG14}, Flowers~\cite{DBLP:conf/icvgip/NilsbackZ08}, Stanford Cars~\cite{DBLP:conf/iccvw/Krause0DF13}, FGVC Aircraft~\cite{DBLP:journals/corr/MajiRKBV13}, Oxford Pets~\cite{DBLP:conf/cvpr/ParkhiVZJ12};
    \item Handwritten digits classification dataset: MNIST~\cite{DBLP:journals/pieee/LeCunBBH98};
    \item Texture classification dataset: DTD~\cite{DBLP:conf/cvpr/CimpoiMKMV14};
    \item Scene classification dataset: SUN397~\cite{DBLP:journals/ijcv/XiaoEHTO16};
    \item Satellite image classification datasets: \\ EuroSAT~\cite{DBLP:journals/staeors/HelberBDB19}, Resisc45~\cite{DBLP:journals/pieee/ChengHL17};
    \item German Traffic Sign Recognition Benchmark: \\ GTSRB~\cite{DBLP:journals/nn/StallkampSSI12};
    \item The metastatic tissue classification dataset: \\ PatchCamelyon (PCAM)~\cite{DBLP:conf/miccai/VeelingLWCW18};
    \item Action Recognition dataset; UCF101~\cite{DBLP:journals/corr/abs-1212-0402};
    \item The CLEVR counting dataset~\cite{DBLP:conf/cvpr/JohnsonHMFZG17};
    \item The Hateful Memes dataset~\cite{DBLP:conf/nips/KielaFMGSRT20};
    \item The Rendered SST2 dataset~\cite{DBLP:conf/icml/RadfordKHRGASAM21};
\end{itemize}
We process these datasets following the open-source code
~\cite{DBLP:conf/icml/GadetskyJB24,DBLP:conf/icml/00030P00024}
~\footnote{https://github.com/XLearning-SCU/2024-ICML-TAC}
~\footnote{https://github.com/mlbio-epfl/turtle}. For CLEVR, we take 2000 random samples as the training set and 500 as the testing set. For the video dataset UCF101, we take the middle frame of each video clip as the input of the pre-trained CLIP vision encoder.

\section{More Implementation Details}
\label{sec:detail}

\subsection{Implementation of the Compared Methods}
\label{sec:other-detail}

\paragraph{K-Means and other traditional clustering methods.} 
We apply k-means clustering~\cite{1967Some} on top of pre-trained features as a simple baseline that only uses knowledge from visual space.
Following the previous work~\cite{DBLP:conf/icml/00030P00024}, we implement traditional clustering methods, such as k-means, using the FAISS~\footnote{https://github.com/facebookresearch/faiss} library for GPU acceleration (\texttt{i.e.}, faiss.Kmeans). Additionally, the relevant parameter settings are consistent with those used in TAC. 
We run the clustering for each dataset 20 times with 300 iterations and keep the best centroids for each iteration ($\mathtt{nredo=20}, \mathtt{niter=300}$). The centroids are L2 normalized after each iteration ($\mathtt{spherical=True}$).

Specifically, for K-means and other traditional clustering methods involved in the ablation studies (\textit{i.e.} Spectral Clustering, Agglomerative Clustering, Bisecting K-Means), we utilized the clustering implementations found in the \texttt{cluster} module of the \texttt{sklearn} library.

\paragraph{Zero-shot CLIP.} To validate the applicability across different scenarios, we use the same prompt list for different datasets, rather than employing a set of prompts manually designed for the characteristics of each dataset (\texttt{dataset to template} in TURTLE). We utilized the same prompt list as in the TAC open-source code, referred to as `the simple ImageNet prompt'~\footnote{https://github.com/openai/CLIP/blob/main/notebooks/\\Prompt\_Engineering\_for\_ImageNet.ipynb}, which consists of seven prompts. The prompts are shown in~\autoref{tab:prompt}.

{\renewcommand{\arraystretch}{1.} 
\begin{table}[h]
    \centering
    \caption{The prompts used for zero-shot CLIP.}
    \begin{tabularx}{0.9\linewidth}{X}
         \hline
         \textbf{Simple ImageNet Templates:} \\
         1. itap of a [class]. \\
         2. a bad photo of the [class]. \\
         3. a origami [class]. \\
         4. a photo of the large [class]. \\
         5. a [class] in a video game. \\
         6. art of the [class]. \\
         7. a photo of the small [class]. \\
         \hline
    \end{tabularx}
    \label{tab:prompt}
\end{table}}

\paragraph{Semantic-Enhanced Image Clustering.} 
SIC~\cite{DBLP:conf/aaai/CaiQ0ZC23} first explores utilizing the knowledge from both visual space and textual space. It attempts to generate pseudo-labels for each image according to the relationships between images and the semantics of the nouns in WordNet.
We utilize the open-source code from SIC~\footnote{https://github.com/Bruce-XJChen/SIC}, employing its default parameters. When extending the datasets used, we build the \textit{dataloader} for each newly added dataset and overwrite the \textit{\_\_getitem\_\_} method accordingly. 

\paragraph{Text-Aided Image Clustering.} 
TAC~\cite{DBLP:conf/icml/00030P00024} also proposes leveraging external knowledge in the textual space. Unlike SIC, TAC computes the text features of each image using noun features, without assigning explicit pseudo-labels.
We utilize the open-source code from TAC~\footnote{https://github.com/XLearning-SCU/2024-ICML-TAC}. We keep the default parameters. We build the \textit{dataloader} for the extra datasets by referencing the \textit{dataloader} in TURTLE.

\paragraph{TURTLE.} 
TURTLE~\cite{DBLP:conf/icml/GadetskyJB24} enables unsupervised transfer from pretrained models to perform image clustering.
It identifies the optimal dataset labeling by maximizing the margins of linear classifiers in the space of single or multiple pretrained models.
It is compatible with any pretrained representations.
We utilize the open-source code from TURTLE~\footnote{https://github.com/mlbio-epfl/turtle}. In the original setup, TURTLE receives image features generated from different pretrained image encoders and trains in multiple spaces. In this paper, we use features from the CLIP image encoder as one space, and the text-enhanced features from TAC and the knowledge-enhanced features from KEC as another space.

{\renewcommand{\arraystretch}{1.0}
\begin{table*}[!h]
    \centering
    \caption{Comparison results across all datasets.}
    \resizebox{0.95\linewidth}{!}{
    \begin{tabular}{@{}lccccccccccccccc@{}}
    \toprule
    \textbf{Dataset} & \multicolumn{3}{c}{\textbf{CIFAR-10}} & \multicolumn{3}{c}{\textbf{CIFAR-100}} & \multicolumn{3}{c}{\textbf{STL-10}} & \multicolumn{3}{c}{\textbf{DTD}} & \multicolumn{3}{c}{\textbf{UCF-101}} \\ \midrule
    \textbf{Metrics} & \textbf{NMI} & \textbf{ACC} & \textbf{ARI} & \textbf{NMI} & \textbf{ACC} & \textbf{ARI} & \textbf{NMI} & \textbf{ACC} & \textbf{ARI} & \textbf{NMI} & \textbf{ACC} & \textbf{ARI} & \textbf{NMI} & \textbf{ACC} & \textbf{ARI} \\ \midrule
    CLIP (k-means) & 73.7 & 80.4 & 64.6 & 59.7 & 43.4 & 29.1 & 91.8 & 94.3 & 89.2 & 58.6 & 45.4 & 29.0 & 80.7 & 59.5 & 51.4 \\
    TAC (no train) & 81.4 & 90.4 & 80.3 & 65.0 & 50.1 & 35.7 & 92.5 & 94.8 & 89.9 & 60.3 & 48.2 & 31.4 & 80.9 & 61.8 & 52.0 \\
    $\mathtt{KEC}$ (no train) & 81.9 & 90.7 & 80.6 & 66.4 & 51.8 & 37.3 & 95.1 & 97.9 & 95.5 & 60.7 & 47.4 & 31.5 & 81.9 & 62.5 & 53.2 \\ \midrule \midrule
    SIC & 84.7 & 92.6 & 84.4 & 63.2 & 47.6 & 34.8 & 95.3 & 98.1 & 95.9 & 59.6 & 45.9 & 30.5 & 81.4 & 65.3 & 56.7 \\
    TAC & 82.9 & 91.5 & 82.3 & 67.5 & 56.1 & 40.8 & 95.6 & 98.2 & 96.1 & 60.8 & 47.8 & 32.4 & 81.3 & 67.4 & 58.2 \\
    $\mathtt{KEC_{TAC}}$ & 84.1 & 92.3 & 84.0 & 68.0 & 57.1 & 41.3 & 95.8 & 98.3 & 96.3 & 62.5 & 51.3 & 36.0 & 81.6 & 67.6 & 58.7 \\ \midrule
    TURTLE (1-space) & 78.6 & 86.5 & 75.1 & 60.8 & 45.0 & 33.1 & 95.8 & 98.4 & 96.4 & 62.9 & 52.9 & 36.7 & 80.9 & 67.1 & 57.1 \\
    $\mathtt{TAC_{TURTLE}}$ & 83.5 & 91.9 & 83.2 & 62.0 & 46.4 & 34.6 & 95.3 & 98.0 & 95.7 & 63.3 & 52.9 & 36.8 & 81.9 & 69.2 & 59.4 \\
    $\mathtt{KEC_{TURTLE}}$ & 83.7 & 92.1 & 83.5 & 62.9 & 47.6 & 35.3 & 96.1 & 98.5 & 96.7 & 63.1 & 52.8 & 36.7 & 82.8 & 70.0 & 60.7 \\ \midrule
    {\color[HTML]{9B9B9B} CLIP (zero-shot)} & 80.7 & 90.0 & 79.3 & 69.9 & 65.0 & 44.7 & 93.8 & 97.0 & 93.7 & 56.1 & 42.6 & 26.6 & 80.3 & 63.7 & 50.2 \\ \toprule
    \textbf{Dataset} & \multicolumn{3}{c}{\textbf{ImageNet}} & \multicolumn{3}{c}{\textbf{Food101}} & \multicolumn{3}{c}{\textbf{SUN397}} & \multicolumn{3}{c}{\textbf{Cars}} & \multicolumn{3}{c}{\textbf{Aircraft}} \\ \midrule
    \textbf{Metrics} & \textbf{NMI} & \textbf{ACC} & \textbf{ARI} & \textbf{NMI} & \textbf{ACC} & \textbf{ARI} & \textbf{NMI} & \textbf{ACC} & \textbf{ARI} & \textbf{NMI} & \textbf{ACC} & \textbf{ARI} & \textbf{NMI} & \textbf{ACC} & \textbf{ARI} \\ \midrule
    CLIP (k-means) & 72.3 & 38.9 & 27.1 & 71.1 & 60.4 & 47.8 & 76.3 & 50.0 & 38.4 & 67.1 & 35.9 & 24.9 & 49.4 & 21.5 & 11.6 \\
    TAC (no train) & 77.5 & 48.4 & 34.5 & 72.5 & 61.5 & 48.1 & 78.6 & 54.4 & 42.5 & 64.7 & 32.6 & 21.3 & 48.6 & 21.6 & 10.7 \\
    $\mathtt{KEC}$ (no train) & 77.7 & 48.6 & 35.5 & 74.6 & 66.2 & 53.3 & 78.2 & 54.0 & 42.3 & 68.3 & 37.6 & 26.6 & 50.6 & 22.5 & 13.3 \\ \midrule \midrule
    SIC & 77.2 & 47.0 & 34.3 & 74.1 & 62.4 & 51.8 & 76.1 & 51.3 & 38.2 & 66.5 & 33.7 & 24.1 & 48.8 & 22.1 & 12.0 \\
    TAC & 78.2 & 54.4 & 39.6 & 74.9 & 68.4 & 54.2 & 74.6 & 41.2 & 32.3 & 60.5 & 25.7 & 16.5 & 44.9 & 19.1 & 9.3 \\
    $\mathtt{KEC_{TAC}}$ & 78.4 & 55.3 & 40.1 & 75.9 & 70.0 & 56.3 & 78.7 & 54.1 & 42.6 & 65.4 & 33.7 & 22.6 & 48.8 & 22.7 & 12.9 \\ \midrule
    TURTLE (1-space) & 66.4 & 25.6 & 15.6 & 72.9 & 64.3 & 52.5 & 77.8 & 54.7 & 42.9 & 69.4 & 41.4 & 30.2 & 49.6 & 24.0 & 13.6 \\
    $\mathtt{TAC_{TURTLE}}$ & 65.6 & 23.8 & 14.5 & 71.5 & 62.2 & 49.7 & 76.9 & 53.3 & 40.5 & 69.4 & 41.5 & 30.2 & 48.8 & 23.8 & 13.0 \\
    $\mathtt{KEC_{TURTLE}}$ & 68.7 & 30.6 & 19.9 & 73.8 & 66.9 & 54.2 & 77.7 & 54.6 & 42.1 & 69.7 & 42.4 & 30.4 & 49.8 & 24.5 & 13.7 \\ \midrule
    {\color[HTML]{9B9B9B} CLIP (zero-shot)} & 81.0 & 63.6 & 45.3 & 82.7 & 83.3 & 69.8 & 80.3 & 64.4 & 47.5 & 77.4 & 59.3 & 43.3 & 48.8 & 22.0 & 11.3 \\ \toprule
    \textbf{Dataset} & \multicolumn{3}{c}{\textbf{Pets}} & \multicolumn{3}{c}{\textbf{Flowers}} & \multicolumn{3}{c}{\textbf{MNIST}} & \multicolumn{3}{c}{\textbf{Eurosat}} & \multicolumn{3}{c}{\textbf{Resisc45}} \\ \midrule
    \textbf{Metrics} & \textbf{NMI} & \textbf{ACC} & \textbf{ARI} & \textbf{NMI} & \textbf{ACC} & \textbf{ARI} & \textbf{NMI} & \textbf{ACC} & \textbf{ARI} & \textbf{NMI} & \textbf{ACC} & \textbf{ARI} & \textbf{NMI} & \textbf{ACC} & \textbf{ARI} \\ \midrule
    CLIP (k-means) & 65.0 & 51.3 & 40.5 & 86.5 & 71.5 & 67.7 & 49.4 & 57.8 & 38.6 & 53.7 & 61.9 & 43.7 & 71.8 & 64.1 & 50.5 \\
    TAC (no train) & 79.2 & 65.8 & 57.8 & 84.5 & 69.4 & 64.8 & 36.9 & 45.5 & 24.7 & 48.8 & 60.9 & 34.5 & 70.7 & 58.3 & 45.6 \\
    $\mathtt{KEC}$ (no train) & 81.2 & 67.8 & 63.3 & 87.3 & 72.8 & 67.5 & 46.9 & 53.7 & 35.2 & 54.6 & 62.7 & 42.8 & 74.3 & 62.4 & 51.5 \\ \midrule \midrule
    SIC & 67.7 & 51.6 & 42.6 & 67.7 & 43.1 & 34.0 & 40.8 & 50.1 & 32.5 & 61.5 & 67.2 & 53.5 & 75.7 & 67.7 & 57.0 \\
    TAC & 83.6 & 77.9 & 69.1 & 80.4 & 64.2 & 59.6 & 42.7 & 54.6 & 34.0 & 53.0 & 67.8 & 45.7 & 72.9 & 68.0 & 53.0 \\
    $\mathtt{KEC_{TAC}}$ & 84.8 & 79.7 & 71.5 & 85.4 & 71.6 & 66.8 & 45.4 & 56.2 & 36.8 & 56.7 & 69.8 & 48.0 & 75.6 & 70.6 & 57.1 \\ \midrule
    TURTLE (1-space) & 71.5 & 60.9 & 48.9 & 90.7 & 87.2 & 79.8 & 42.8 & 49.6 & 34.4 & 57.9 & 63.6 & 48.2 & 75.3 & 70.9 & 57.0 \\
    $\mathtt{TAC_{TURTLE}}$ & 73.7 & 63.4 & 52.5 & 90.8 & 86.8 & 79.1 & 41.5 & 50.7 & 33.6 & 60.0 & 67.7 & 51.8 & 74.4 & 69.4 & 55.4 \\
    $\mathtt{KEC_{TURTLE}}$ & 81.0 & 72.4 & 63.3 & 92.4 & 88.4 & 82.7 & 47.8 & 57.8 & 39.4 & 63.0 & 76.8 & 57.4 & 76.4 & 73.6 & 59.3 \\ \midrule
    {\color[HTML]{9B9B9B} CLIP (zero-shot)} & 87.6 & 84.9 & 75.4 & 79.4 & 67.5 & 58.7 & 28.4 & 38.1 & 11.9 & 41.0 & 45.6 & 26.4 & 64.4 & 53.4 & 35.3 \\ \toprule
    \textbf{Dataset} & \multicolumn{3}{c}{\textbf{GTSRB}} & \multicolumn{3}{c}{\textbf{PCAM}} & \multicolumn{3}{c}{\textbf{CLEVR}} & \multicolumn{3}{c}{\textbf{HatefulMemes}} & \multicolumn{3}{c}{\textbf{SST}} \\ \midrule
    \textbf{Metrics} & \textbf{NMI} & \textbf{ACC} & \textbf{ARI} & \textbf{NMI} & \textbf{ACC} & \textbf{ARI} & \textbf{NMI} & \textbf{ACC} & \textbf{ARI} & \textbf{NMI} & \textbf{ACC} & \textbf{ARI} & \textbf{NMI} & \textbf{ACC} & \textbf{ARI} \\ \midrule
    CLIP (k-means) & 52.5 & 32.1 & 23.1 & 10.2 & 63.3 & 7.0 & 18.9 & 27.6 & 8.1 & 2.0 & 58.2 & 2.4 & 0.4 & 53.7 & 0.5 \\
    TAC (no train) & 48.3 & 31.0 & 20.2 & 1.5 & 56.7 & 1.8 & 11.8 & 23.6 & 5.4 & 1.2 & 56.4 & 1.4 & 0.1 & 51.9 & 0.1 \\
    $\mathtt{KEC}$ (no train) & 51.9 & 32.6 & 22.6 & 10.0 & 63.6 & 7.4 & 16.3 & 26.6 & 8.9 & 1.6 & 57.4 & 2.0 & 0.5 & 53.8 & 0.6 \\ \midrule \midrule
    SIC & 52.3 & 36.4 & 28.3 & 0.0 & 51.1 & 0.0 & 5.2 & 20.6 & 1.5 & 0.1 & 51.8 & 0.0 & 0.5 & 54.1 & 0.6 \\
    TAC & 40.5 & 26.7 & 16.8 & 0.0 & 50.3 & 0.0 & 8.4 & 22.4 & 3.6 & 0.6 & 54.6 & 0.6 & 0.1 & 52.0 & 0.1 \\
    $\mathtt{KEC_{TAC}}$ & 48.6 & 29.6 & 20.1 & 9.0 & 63.2 & 7.1 & 15.9 & 25.4 & 6.8 & 2.3 & 59.0 & 3.0 & 0.9 & 55.5 & 1.1 \\ \midrule
    TURTLE (1-space) & 51.1 & 33.9 & 25.1 & 0.0 & 51.3 & 0.0 & 16.0 & 24.6 & 7.0 & 1.1 & 55.6 & 1.8 & 0.2 & 52.4 & 0.4 \\
    $\mathtt{TAC_{TURTLE}}$ & 48.1 & 29.9 & 22.1 & 0.0 & 50.3 & 0.0 & 16.6 & 25.8 & 7.5 & 1.4 & 56.0 & 2.0 & 0.2 & 52.4 & 0.5 \\
    $\mathtt{KEC_{TURTLE}}$ & 48.8 & 32.4 & 23.5 & 8.9 & 62.7 & 6.8 & 18.8 & 28.6 & 8.0 & 2.0 & 58.4 & 2.6 & 0.8 & 55.2 & 1.0 \\ \midrule
    {\color[HTML]{9B9B9B} CLIP (zero-shot)} & 46.2 & 32.3 & 22.3 & 3.1 & 52.1 & 0.2 & 16.4 & 4.4 & 8.0 & 0.5 & 53.4 & 0.3 & 0.0 & 51.0 & 0.0 \\ \bottomrule
    \end{tabular}}
    \label{tab:main_all_dataset}
\end{table*}}

\subsection{Implementation of the Proposed Method}
\label{sec:our-detail}

We precompute the features of the input images for all datasets and the features of the nouns from WordNet before all the experiments, using the \texttt{batch\_size} of 8192.

Code and more implementation details will be available.
The comparison methods are integrated into our project using their original code. Our approach offers good adaptability and can quickly be applied to different clustering strategies via configuration file modifications. 

In all experiments, we extract knowledge from \texttt{GPT-4o} and set the \texttt{temperature} to 0.1. Specifically, we utilize the \texttt{AsyncOpenAI} interface for asynchronous processing to improve efficiency, setting \texttt{max\_concurrent} to 20 within Python's \texttt{async} framework.

\section{Main Results Across All Datasets}
\label{sec:all}

We present the complete numerical results in~\autoref{tab:main_all_dataset}. CLIP (k-means) and TURTLE (1-space) utilize only visual space knowledge, while zero-shot CLIP incorporates ground-truth label knowledge from the textual space. SIC, TAC (no train), and KEC utilize textual space knowledge differently. Additionally, we apply two training strategies for the multi-space knowledge: TAC and TURTLE.

It is worth emphasizing again that, compared to KEC and other methods, CLIP (zero-shot) leverages additional fine-grained information, \textit{i.e.} the ground-truth labels for each image (\textit{e.g.}, `Boeing 747', `Audi A4'), which effectively simplifies the task setting. Therefore, CLIP (zero-shot) could be considered as a loosely constrained upper bound on performance. 
Our proposed method constructs discriminative textual knowledge and achieves superior performance over existing baselines across a wide range of datasets. In many cases, it even surpasses CLIP (zero-shot), further demonstrating the effectiveness and generality of our method.

We also observe that the extent of performance improvement of KEC varies across different datasets.\\ 
1. For higher-level or subjective categories, the improvements are less pronounced compared to other datasets. This suggests that such tasks may require a certain degree of human-machine collaboration to better construct effective textual knowledge (which will be further discussed in the Limitations and Future Work section).\\
2. For some fine-grained datasets, such as Aircraft and Cars, which focus on specific domains, CLIP (zero-shot) benefits from directly using ground-truth fine-grained labels, enabling it to effortlessly focus on subtle category differences. In contrast, without additional constraints, KEC adopts a more general perspective to interpret these categories, demonstrating strong generalization capability. Consequently, KEC achieves only modest performance gains or may slightly underperform compared to CLIP zero-shot, yet it still surpasses other baseline methods.

{\renewcommand{\arraystretch}{1.}
\begin{table*}[!ht]
\centering
\caption{Ablation results across all datasets. \textit{Con.} and \textit{Des.} represent the name and its description of a concept, respectively. \textit{UA} and \textit{BA} represent the uni-concept attribute and bi-concept attribute.}
\resizebox{\linewidth}{!}{
\begin{tabular}{@{}cccc|ccc|ccc|ccc|ccc|ccc@{}}
\toprule
\multirow{2}{*}{\textbf{Con.}} & \multirow{2}{*}{\textbf{Des.}} & \multirow{2}{*}{\textbf{UA}} & \multirow{2}{*}{\textbf{BA}} & \multicolumn{3}{c|}{\textbf{CIFAR-10}} & \multicolumn{3}{c|}{\textbf{CIFAR-100}} & \multicolumn{3}{c|}{\textbf{STL-10}} & \multicolumn{3}{c|}{\textbf{DTD}} & \multicolumn{3}{c}{\textbf{UCF-101}} \\ \cmidrule(l){5-19} 
 &  &  &  & \textbf{NMI} & \textbf{ACC} & \textbf{ARI} & \textbf{NMI} & \textbf{ACC} & \textbf{ARI} & \textbf{NMI} & \textbf{ACC} & \textbf{ARI} & \textbf{NMI} & \textbf{ACC} & \textbf{ARI} & \textbf{NMI} & \textbf{ACC} & \textbf{ARI} \\ \midrule
\checkmark &  &  &  & 80.9 & 89.7 & 79.7 & 64.5 & 49.4 & 36.0 & 94.0 & 97.0 & 94.5 & 58.0 & 43.6 & 26.4 & 80.5 & 60.2 & 51.8 \\
 & \checkmark &  &  & 80.1 & 89.0 & 78.3 & 64.2 & 47.9 & 33.4 & 93.7 & 96.7 & 94.0 & 57.4 & 46.3 & 28.8 & 80.7 & 60.7 & 52.7 \\
\checkmark & \checkmark &  &  & 81.4 & 90.1 & 80.5 & 64.6 & 49.1 & 36.0 & 92.4 & 96.5 & 94.4 & 58.2 & 46.8 & 28.7 & 80.7 & 61.2 & 51.3 \\ \midrule
\checkmark & \checkmark & \checkmark & \multicolumn{1}{l|}{} & 81.9 & 90.5 & 80.2 & 66.0 & 50.3 & 36.6 & 95.1 & 97.9 & 95.5 & 59.8 & 46.2 & 30.1 & 81.9 & 60.8 & 52.0 \\
\checkmark & \checkmark & \multicolumn{1}{l}{} & \checkmark & 81.9 & 90.6 & 80.3 & 65.7 & 50.9 & 36.4 & 95.1 & 97.9 & 95.5 & 60.2 & 46.8 & 31.1 & 81.7 & 61.8 & 53.1 \\
\checkmark & \checkmark & \checkmark & \checkmark & 81.9 & 90.7 & 80.6 & 66.4 & 51.8 & 37.3 & 95.1 & 97.9 & 95.5 & 60.7 & 47.4 & 31.5 & 81.9 & 62.5 & 53.2 \\ \toprule
\multirow{2}{*}{\textbf{Con.}} & \multirow{2}{*}{\textbf{Des.}} & \multirow{2}{*}{\textbf{UA}} & \multirow{2}{*}{\textbf{BA}} & \multicolumn{3}{c|}{\textbf{ImageNet}} & \multicolumn{3}{c|}{\textbf{Food101}} & \multicolumn{3}{c|}{\textbf{SUN397}} & \multicolumn{3}{c|}{\textbf{Cars}} & \multicolumn{3}{c}{\textbf{Aircraft}} \\ \cmidrule(l){5-19} 
 &  &  &  & \textbf{NMI} & \textbf{ACC} & \textbf{ARI} & \textbf{NMI} & \textbf{ACC} & \textbf{ARI} & \textbf{NMI} & \textbf{ACC} & \textbf{ARI} & \textbf{NMI} & \textbf{ACC} & \textbf{ARI} & \textbf{NMI} & \textbf{ACC} & \textbf{ARI} \\ \midrule
\checkmark &  &  &  & 75.5 & 45.3 & 33.4 & 73.7 & 65.7 & 52.5 & 77.2 & 52.4 & 40.2 & 67.6 & 36.5 & 24.3 & 49.5 & 22.5 & 12.5 \\
 & \checkmark &  &  & 74.7 & 45.2 & 33.0 & 73.2 & 63.2 & 50.8 & 77.0 & 52.6 & 40.6 & 66.0 & 35.7 & 23.8 & 48.9 & 21.4 & 10.8 \\
\checkmark & \checkmark &  &  & 75.8 & 45.3 & 33.1 & 73.9 & 66.0 & 52.4 & 77.6 & 53.0 & 41.6 & 68.3 & 36.4 & 25.8 & 49.1 & 22.5 & 12.2 \\ \midrule
\checkmark & \checkmark & \checkmark & \multicolumn{1}{l|}{} & 76.9 & 46.8 & 35.1 & 74.7 & 66.2 & 53.5 & 78.2 & 53.8 & 42.3 & 68.3 & 37.5 & 26.4 & 50.4 & 22.4 & 13.1 \\
\checkmark & \checkmark & \multicolumn{1}{l}{} & \checkmark & 77.0 & 47.0 & 35.5 & 74.8 & 66.7 & 53.7 & 78.2 & 53.7 & 42.5 & 68.2 & 37.8 & 26.6 & 50.6 & 22.7 & 13.2 \\
\checkmark & \checkmark & \checkmark & \checkmark & 77.7 & 48.6 & 35.5 & 74.6 & 66.2 & 53.3 & 78.2 & 54.0 & 42.3 & 68.3 & 37.6 & 26.6 & 50.6 & 22.5 & 13.3 \\ \toprule
\multirow{2}{*}{\textbf{Con.}} & \multirow{2}{*}{\textbf{Des.}} & \multirow{2}{*}{\textbf{UA}} & \multirow{2}{*}{\textbf{BA}} & \multicolumn{3}{c|}{\textbf{Pets}} & \multicolumn{3}{c|}{\textbf{Flowers}} & \multicolumn{3}{c|}{\textbf{MNIST}} & \multicolumn{3}{c|}{\textbf{Eurosat}} & \multicolumn{3}{c}{\textbf{Resisc45}} \\ \cmidrule(l){5-19} 
 &  &  &  & \textbf{NMI} & \textbf{ACC} & \textbf{ARI} & \textbf{NMI} & \textbf{ACC} & \textbf{ARI} & \textbf{NMI} & \textbf{ACC} & \textbf{ARI} & \textbf{NMI} & \textbf{ACC} & \textbf{ARI} & \textbf{NMI} & \textbf{ACC} & \textbf{ARI} \\ \midrule
\checkmark &  &  &  & 78.9 & 66.0 & 61.1 & 84.9 & 68.9 & 62.3 & 45.4 & 50.6 & 32.7 & 52.2 & 61.1 & 41.7 & 71.7 & 63.8 & 50.6 \\
 & \checkmark &  &  & 78.9 & 64.9 & 60.9 & 85.0 & 68.6 & 62.0 & 45.6 & 50.9 & 31.9 & 51.9 & 60.9 & 40.4 & 72.1 & 64.2 & 50.6 \\
\checkmark & \checkmark &  &  & 79.0 & 64.7 & 60.5 & 85.5 & 70.8 & 63.6 & 45.9 & 51.7 & 32.6 & 52.0 & 62.0 & 41.5 & 72.2 & 61.8 & 51.2 \\ \midrule
\checkmark & \checkmark & \checkmark & \multicolumn{1}{l|}{} & 80.5 & 67.3 & 63.1 & 87.3 & 72.6 & 67.6 & 46.7 & 53.6 & 35.0 & 54.4 & 62.3 & 42.4 & 74.3 & 63.2 & 52.3 \\
\checkmark & \checkmark & \multicolumn{1}{l}{} & \checkmark & 80.2 & 66.7 & 62.8 & 86.8 & 72.2 & 65.2 & 45.4 & 49.0 & 29.5 & 54.5 & 62.4 & 42.4 & 74.6 & 62.8 & 52.0 \\
\checkmark & \checkmark & \checkmark & \checkmark & 81.2 & 67.8 & 63.3 & 87.3 & 72.5 & 67.5 & 46.9 & 53.7 & 35.2 & 54.6 & 62.7 & 42.8 & 74.3 & 62.4 & 51.5 \\ \toprule
\multirow{2}{*}{\textbf{Con.}} & \multirow{2}{*}{\textbf{Des.}} & \multirow{2}{*}{\textbf{UA}} & \multirow{2}{*}{\textbf{BA}} & \multicolumn{3}{c|}{\textbf{GTSRB}} & \multicolumn{3}{c|}{\textbf{PCAM}} & \multicolumn{3}{c|}{\textbf{CLEVR}} & \multicolumn{3}{c|}{\textbf{HatefulMemes}} & \multicolumn{3}{c}{\textbf{SST}} \\ \cmidrule(l){5-19} 
 &  &  &  & \textbf{NMI} & \textbf{ACC} & \textbf{ARI} & \textbf{NMI} & \textbf{ACC} & \textbf{ARI} & \textbf{NMI} & \textbf{ACC} & \textbf{ARI} & \textbf{NMI} & \textbf{ACC} & \textbf{ARI} & \textbf{NMI} & \textbf{ACC} & \textbf{ARI} \\ \midrule
\checkmark &  &  &  & 48.9 & 28.6 & 21.2 & 9.0 & 62.6 & 6.4 & 14.2 & 23.2 & 5.5 & 1.6 & 57.2 & 1.9 & 0.5 & 53.9 & 0.6 \\
 & \checkmark &  &  & 49.6 & 29.7 & 20.5 & 8.1 & 62.7 & 6.5 & 15.4 & 24.2 & 6.3 & 1.4 & 56.8 & 1.7 & 0.6 & 54.3 & 0.7 \\
\checkmark & \checkmark &  &  & 49.9 & 30.1 & 21.4 & 9.0 & 62.6 & 6.4 & 15.8 & 25.8 & 8.0 & 1.6 & 57.2 & 1.9 & 0.6 & 54.1 & 0.6 \\ \midrule
\checkmark & \checkmark & \checkmark & \multicolumn{1}{l|}{} & 51.8 & 32.4 & 22.5 & 9.9 & 63.6 & 7.4 & 16.0 & 26.6 & 8.6 & 1.6 & 57.4 & 2.0 & 0.5 & 54.0 & 0.6 \\
\checkmark & \checkmark & \multicolumn{1}{l}{} & \checkmark & 51.9 & 32.4 & 22.7 & 10.0 & 63.7 & 7.5 & 16.3 & 26.6 & 8.9 & 1.7 & 57.6 & 2.1 & 0.5 & 53.8 & 0.5 \\
\checkmark & \checkmark & \checkmark & \checkmark & 51.9 & 32.6 & 22.6 & 10.0 & 63.6 & 7.4 & 16.3 & 26.6 & 8.9 & 1.6 & 57.4 & 2.0 & 0.5 & 53.8 & 0.6 \\ \bottomrule
\end{tabular}}
\label{tab:ab_all_dataset}
\end{table*}}

As previously mentioned, KEC possesses the ability to construct customized knowledge (by human-machine collaboration). To verify this, we conduct a simple experiment: assuming the task is to distinguish specific types of cars or aircraft, we guide KEC to selectively acquire concepts and attributes related to cars and aircraft and build hierarchical textual knowledge, which we refer to as KEC$^+$. As shown in~\autoref{tab:kec+}. Without access to ground-truth labels, KEC$^+$ achieves further performance improvements through simple human-machine interaction and customization.

{\renewcommand{\arraystretch}{1}
\begin{table}[t]
\centering
\caption{Results of KEC with targeted orientation.}
\resizebox{\linewidth}{!}{
\begin{tabular}{@{}l|cccccc@{}}
\midrule
\multirow{2}{*}{\textbf{Methods}} & \multicolumn{3}{c}{\textbf{Aircraft}} & \multicolumn{3}{c}{\textbf{Cars}} \\
 & \textbf{NMI} & \textbf{ACC} & \textbf{ARI} & \textbf{NMI} & \textbf{ACC} & \textbf{ARI} \\ \midrule
\textbf{CLIP (zero-shot)} & 48.8 & 22.0 & 10.7 & 77.4 & 59.3 & 43.3 \\
\textbf{KEC} & 50.6 & 22.5 & 13.3 & 68.3 & 37.6 & 26.6 \\
\textbf{KEC$^+$} & 52.1 & 24.2 & 15.8 & 74.8 & 53.7 & 42.6 \\
\midrule
\end{tabular}}
\label{tab:kec+}
\end{table}}

\section{More Ablation Studies}
\label{sec:abl}

\subsection{Results of ablation studies across all datasets}

We present the results of ablation experiments for the proposed method across each dataset in~\autoref{tab:ab_all_dataset}. We gradually introduce different levels of knowledge and their components using the LLMs.

\subsection{Further analysis on parameter selections}

\paragraph{Parameters in Image-Text Mapping.} 

During the construction of image-text mappings, we aim to associate each image with its relevant semantics (nouns in WordNet). To preserve a strong generalization capability, we intentionally avoid meticulous parameter tuning at this stage. Our method aims to distill hierarchical and discriminative knowledge from the initial textual knowledge, which contains considerable semantic redundancy. Following the settings in TAC, we set the number of clusters as $N_v/300$ (where $N_v$ denotes the number of images), and select the TopK=5 most relevant nouns for each image.

To evaluate the sensitivity of the mapping stage in KEC to these parameters, we systematically varied them and conducted corresponding experiments. The results, shown in~\autoref{tab:top} and~\autoref{tab:nv}, indicate that KEC exhibits strong robustness under different configurations (\textit{i.e.}, textual knowledge generated with varying degrees of semantic redundancy through different mapping strategies). Although different parameter settings may cause slight performance variations across datasets, the overall impact remains limited. Therefore, meticulously tuning parameters for each dataset is labor-intensive, marginally beneficial, and ultimately unnecessary. This further demonstrates the strong generalization capability of KEC across diverse datasets.

{\renewcommand{\arraystretch}{1.2}
\begin{table}[t]
\centering
\caption{Results for different numbers of selected nouns.}
\resizebox{\linewidth}{!}{
\begin{tabular}{@{}c|ccccccccc@{}}
\toprule
\multirow{2}{*}{\textbf{TopK}} & \multicolumn{3}{c}{\textbf{Flowers}}       & \multicolumn{3}{c}{\textbf{Pets}}          & \multicolumn{3}{c}{\textbf{CLEVR}}         \\
                               & \textbf{NMI} & \textbf{ACC} & \textbf{ARI} & \textbf{NMI} & \textbf{ACC} & \textbf{ARI} & \textbf{NMI} & \textbf{ACC} & \textbf{ARI} \\ \midrule
\textbf{1}                     & 86.9         & 72.1         & 69.2         & 80.1         & 67.9         & 62.1         & 17.0         & 27.0         & 7.8          \\
\textbf{3}                     & 87.0         & 72.6         & 66.9         & 80.0         & 68.2         & 62.7         & 16.0         & 26.6         & 8.6          \\
\rowcolor[HTML]{DDEDF1} 
\textbf{5}                     & 87.3         & 72.8         & 67.5         & 81.2         & 67.8         & 63.3         & 16.3         & 26.6         & 8.9          \\
\textbf{10}                    & 87.7         & 73.6         & 68.2         & 80.0         & 67.5         & 62.6         & 16.1         & 26.4         & 8.7          \\ \bottomrule
\end{tabular}}
\label{tab:top}
\end{table}}

{\renewcommand{\arraystretch}{1.2}
\begin{table}[t]
\centering
\caption{Results for different cluster numbers in the initial Image-Text Mapping.}
\resizebox{\linewidth}{!}{
\begin{tabular}{@{}c|ccccccccc@{}}
\toprule
                            & \multicolumn{3}{c}{\textbf{Flowers}}       & \multicolumn{3}{c}{\textbf{Pets}}          & \multicolumn{3}{c}{\textbf{CLEVR}}         \\
\multirow{-2}{*}{\textbf{}} & \textbf{NMI} & \textbf{ACC} & \textbf{ARI} & \textbf{NMI} & \textbf{ACC} & \textbf{ARI} & \textbf{NMI} & \textbf{ACC} & \textbf{ARI} \\ \midrule
\textbf{$N_v/50$}           & 87.0         & 72.5         & 67.2         & 81.3         & 67.9         & 63.5         & 16.5         & 26.7         & 8.9          \\
\textbf{$N_v/100$}         & 87.4         & 72.7         & 67.3         & 81.2         & 67.8         & 63.2         & 16.3         & 26.7         & 8.9          \\
\rowcolor[HTML]{DDEDF1} 
\textbf{$N_v/300$}          & 87.3         & 72.8         & 67.5         & 81.2         & 67.8         & 63.3         & 16.3         & 26.6         & 8.9          \\
\textbf{$N_v/500$}          & 87.6         & 73.0         & 67.8         & 81.1         & 67.5         & 62.9         & 15.8         & 26.1         & 8.1          \\ \bottomrule
\end{tabular}}
\label{tab:nv}
\end{table}}

\paragraph{Weight to balance visual and textual similarity.}

To construct multi-modal similarity, we analyze the impact of the fusion weight $\alpha$ in Equation~2. Specifically, we vary $\alpha$ from 1.0 to 0.0, gradually increasing the contribution of textual knowledge in the similarity computation. The results, as shown in~\autoref{tab:alpha}, lead to the following observations:

\begin{itemize}
    \item When $\alpha=1.0$ (\textit{i.e.}, only using visual similarity), it fails to leverage any textual knowledge. As a result, the constructed text space becomes nearly identical to the visual space, yielding performance comparable to the baseline without improvement.
    \item As $\alpha$ decreases, performance initially improves and then declines, indicating that at this stage, textual knowledge relies on the guidance of visual knowledge to extract more discriminative concepts. When textual knowledge dominates, redundant or mismatched granularity of extracted nouns (either too fine-grained or overly broad) tends to disrupt semantic distillation, resulting in degraded performance. Notably, when $\alpha=0$, \textit{i.e.} relying entirely on textual knowledge, the results are the worst.
    \item To further determine the optimal setting, we conduct additional experiments with $\alpha=0.7$ and $0.9$ around $0.8$. Considering the trade-off between performance and generalization across datasets, we empirically set $\alpha=0.8$ as the default value.
\end{itemize}

{\renewcommand{\arraystretch}{1.2}
\begin{table}[t]
\centering
\caption{Performance analysis based on varying the weight value of visual and textual similarity. The \textbf{bold} numbers indicate the best results. In practice, we set $\alpha$ to 0.8.}
\resizebox{\linewidth}{!}{
\begin{tabular}{@{}c|ccccccccc@{}}
\toprule
                                    & \multicolumn{3}{c}{\textbf{Flowers}}          & \multicolumn{3}{c}{\textbf{Pets}}             & \multicolumn{3}{c}{\textbf{CLEVR}}           \\
\multirow{-2}{*}{\textbf{$\alpha$}} & \textbf{NMI}  & \textbf{ACC}  & \textbf{ARI}  & \textbf{NMI}  & \textbf{ACC}  & \textbf{ARI}  & \textbf{NMI}  & \textbf{ACC}  & \textbf{ARI} \\ \midrule
\textbf{1.0}                        & 86.5          & 72.9          & 67.9          & 78.2          & 65.3          & 58.6          & 15.5          & 25.6          & 6.9          \\ \midrule
\underline{0.9}                           & \textbf{87.8} & \textbf{73.1} & \textbf{68.2} & 81.0          & 67.7          & 62.9          & 15.1          & 25.4          & 6.7          \\
\rowcolor[HTML]{DDEDF1} 
\textbf{0.8}                        & 87.3          & 72.8          & 67.5          & \textbf{81.2} & \textbf{67.8} & \textbf{63.3} & \textbf{16.3} & \textbf{26.6} & \textbf{8.9} \\
\underline{0.7}                           & 86.5          & 72.9          & 67.9          & 80.5          & 67.0          & 62.7          & 16.0          & 26.6          & 8.5          \\ \midrule
\textbf{0.6}                        & 84.0          & 67.7          & 64.5          & 75.5          & 62.3          & 55.1          & 15.1          & 25.8          & 8.0          \\
\textbf{0.4}                        & 76.5          & 62.7          & 59.7          & 70.5          & 56.0          & 47.8          & 14.7          & 24.8          & 6.7          \\
\textbf{0.2}                        & 70.6          & 57.9          & 52.2          & 66.5          & 50.6          & 42.4          & 13.1          & 24.2          & 6.3          \\
\textbf{0.0}                        & 69.5          & 56.4          & 50.7          & 65.5          & 50.0          & 41.7          & 11.5          & 22.8          & 4.5          \\ \bottomrule
\end{tabular}}
\label{tab:alpha}
\end{table}}

\paragraph{Threshold in knowledge construction.}

For threshold selection, we conduct experiments by gradually lowering the threshold from 0.9. The results are shown in~\autoref{tab:beta}. As the threshold decreases, more concepts and attributes are extracted, which significantly increases computational cost and time consumption. However, this does not lead to continuous performance improvement and may even result in performance degradation under certain settings. Considering the trade-off between performance and computational efficiency across different datasets, we empirically set the threshold to 0.8 as the default value.

{\renewcommand{\arraystretch}{1.2}
\begin{table}[t]
\centering
\caption{Performance analysis based on varying the threshold value used in KEC. The \textbf{bold} numbers indicate the best results. The threshold $\beta$ is empirically set to 0.8.}
\resizebox{\linewidth}{!}{
\begin{tabular}{c|ccccccccc}
\hline
                                     & \multicolumn{3}{c}{\textbf{Flowers}}          & \multicolumn{3}{c}{\textbf{Pets}}             & \multicolumn{3}{c}{\textbf{CLEVR}}           \\
\multirow{-2}{*}{\textbf{$\beta$}}   & \textbf{NMI}  & \textbf{ACC}  & \textbf{ARI}  & \textbf{NMI}  & \textbf{ACC}  & \textbf{ARI}  & \textbf{NMI}  & \textbf{ACC}  & \textbf{ARI} \\ \hline
0.9                                  & 86.7          & 71.4          & 65.9          & 79.9          & 64.9          & 60.7          & 15.4          & 24.8          & 6.7          \\
\rowcolor[HTML]{DDEDF1} 
\cellcolor[HTML]{DDEDF1}\textbf{0.8} & 87.3          & \textbf{72.8} & \textbf{67.5} & \textbf{81.2} & 67.8          & \textbf{63.3} & \textbf{16.3} & \textbf{26.6} & \textbf{8.9} \\
0.7                                  & \textbf{87.5} & 71.7          & 67.0          & 80.0          & \textbf{68.8} & 62.8          & 14.7          & 25.4          & 6.5          \\
0.6                                  & 86.3          & 71.2          & 64.9          & 80.5          & 67.2          & 62.1          & 14.5          & 24.2          & 5.9          \\
0.5                                  & 83.5          & 67.2          & 63.8          & 78.8          & 66.1          & 60.9          & 14.0          & 24.4          & 6.1          \\ \hline
\end{tabular}}
\label{tab:beta}
\end{table}}

\subsection{Compatibility with traditional methods}

To demonstrate the compatibility of our method, we evaluate KEC using three additional traditional clustering algorithms beyond K-Means: spectral clustering, agglomeration clustering, and bisecting K-Means. 
All experiments are conducted on the concatenated features $[\mathbf{x}_i;\kappa_i]$, with the number of clusters set to match the ground-truth class count.
As shown in~\autoref{tab:var}, KEC consistently achieves strong performance across all clustering algorithms. These results highlight that the knowledge-enhanced features produced by KEC are broadly compatible with a range of standard clustering algorithms, offering flexibility and ease of integration into diverse practical scenarios.

{\renewcommand{\arraystretch}{1.2}
\begin{table}[t]
\centering
\caption{Clustering performance of KEC using different downstream traditional clustering algorithms.}
\resizebox{2.7in}{!}{
\begin{tabular}{@{}llll@{}}
\midrule
 & \multicolumn{3}{c}{\textbf{Average}} \\
 & \multicolumn{1}{c}{\textbf{NMI}} & \multicolumn{1}{c}{\textbf{ACC}} & \multicolumn{1}{c}{\textbf{ARI}} \\ \midrule
\textbf{K-Means} & \underline{58.0} & \underline{56.6} & \textbf{38.5} \\
\textbf{Spectral Clustering} & 57.0 & 55.8 & \underline{38.4} \\
\textbf{Agglomerative Clustering} & \textbf{58.2} & \textbf{56.9} & 38.2 \\
\textbf{Bisecting K-Means} & 57.2 & 56.2 & 37.6 \\ \midrule
\end{tabular}}
\label{tab:var}
\end{table}}

{\renewcommand{\arraystretch}{1}
\begin{table*}[t]
\centering
\caption{Performance of KEC under different pretrained visual-language models. We report average results across 20 datasets, along with detailed scores on three representative datasets. \textbf{Bold} numbers indicate the best performance within each setting.}
\resizebox{\linewidth}{!}{
\begin{tabular}{@{}ll|lllccccccccc@{}}
\midrule
\multirow{2}{*}{\textbf{Model}} & \multirow{2}{*}{\textbf{Method}} & \multicolumn{3}{c}{\textbf{Average}} & \multicolumn{3}{c}{\textbf{Flowers}} & \multicolumn{3}{c}{\textbf{Pets}} & \multicolumn{3}{c}{\textbf{CLEVR}} \\
 &  & \multicolumn{1}{c}{\textbf{NMI}} & \multicolumn{1}{c}{\textbf{ACC}} & \multicolumn{1}{c}{\textbf{ARI}} & \textbf{NMI} & \textbf{ACC} & \textbf{ARI} & \textbf{NMI} & \textbf{ACC} & \textbf{ARI} & \textbf{NMI} & \textbf{ACC} & \textbf{ARI} \\ \midrule
\multirow{3}{*}{\textbf{CLIP ResNet-50}} & \textbf{CLIP (K-Means)} & 50.5 & 47.9 & 28.4 & 83.6 & 70.4 & 63.6 & 55.9 & 38.8 & 27.5 & \textbf{7.9} & 20.2 & \textbf{2.5} \\
 & \textbf{TAC (no train)} & 50.9 & 48.4 & 29.1 & 83.3 & 67.0 & 60.4 & 76.0 & 60.1 & 52.1 & 6.8 & 20.0 & 1.8 \\
 & \textbf{KEC (no train)} & \textbf{51.9} & \textbf{49.3} & \textbf{30.5} & \textbf{84.1} & \textbf{71.1} & \textbf{64.2} & \textbf{78.6} & \textbf{67.9} & \textbf{58.1} & 7.1 & \textbf{20.6} & 2.0 \\ \midrule
\multirow{3}{*}{\textbf{CLIP ViT-B/16}} & \textbf{CLIP (K-Means)} & 58.1 & 56.0 & 38.0 & 86.1 & 72.1 & 68.4 & 69.1 & 53.3 & 44.8 & 12.6 & 26.0 & 5.8 \\
 & \textbf{TAC (no train)} & 56.5 & 54.2 & 36.5 & 86.2 & 66.1 & 62.3 & 83.1 & 70.2 & 64.0 & 9.7 & 23.2 & 4.0 \\
 & \textbf{KEC (no train)} & \textbf{60.0} & \textbf{59.4} & \textbf{41.6} & \textbf{90.1} & \textbf{75.9} & \textbf{72.0} & \textbf{85.2} & \textbf{77.0} & \textbf{69.8} & \textbf{14.5} & \textbf{26.2} & \textbf{6.8} \\ \midrule
\multirow{3}{*}{\textbf{ImageBind-Huge}} & \textbf{CLIP (K-Means)} & 61.0 & 58.8 & 42.2 & \textbf{93.8} & 72.2 & \textbf{72.9} & 85.2 & 70.6 & 67.3 & 16.4 & 25.8 & \textbf{9.5} \\
 & \textbf{TAC (no train)} & 62.1 & 59.6 & 43.0 & 89.3 & 70.9 & 70.1 & 85.7 & 68.9 & 65.6 & 15.3 & 24.6 & 7.2 \\
 & \textbf{KEC (no train)} & \textbf{64.8} & \textbf{62.7} & \textbf{47.5} & 92.0 & \textbf{72.8} & \textbf{72.9} & \textbf{88.4} & \textbf{73.1} & \textbf{71.4} & \textbf{18.2} & \textbf{26.2} & 9.2 \\ \midrule
\end{tabular}}
\label{tab:backbone}
\end{table*}}

{\renewcommand{\arraystretch}{1.}
\begin{table*}[t]
\centering
\caption{Performance comparison of KEC with different LLMs for knowledge construction. 
The LLMs are from different providers and with various model sizes. The results demonstrate the robustness of KEC.}
\resizebox{4.7in}{!}{
\begin{tabular}{@{}l|ccccccccc@{}}
\midrule
\multirow{2}{*}{\textbf{LLMs}} & \multicolumn{3}{c}{\textbf{Flowers}} & \multicolumn{3}{c}{\textbf{Pets}} & \multicolumn{3}{c}{\textbf{CLEVR}} \\
 & \textbf{NMI} & \textbf{ACC} & \textbf{ARI} & \textbf{NMI} & \textbf{ACC} & \textbf{ARI} & \textbf{NMI} & \textbf{ACC} & \textbf{ARI} \\ \midrule
\textbf{Claude-3-7-Sonnet} & 86.5 & 72.9 & 67.9 & 79.8 & 67.1 & 59.6 & 16.7 & 26.8 & 8.6 \\
\textbf{Gemini-2.0-Flash} & 86.5 & 72.9 & 67.9 & 81.0 & 67.8 & 62.7 & 17.0 & 26.8 & 9.0 \\
\textbf{Deepseek-V3} & 80.1 & 65.9 & 60.6 & 86.0 & 72.7 & 66.0 & 16.1 & 26.0 & 8.1 \\
\textbf{Qwen-Turbo} & 80.6 & 70.8 & 62.4 & 86.1 & 70.3 & 64.9 & 16.0 & 27.4 & 8.8 \\
\textbf{Qwen2.5-7B-Instruct} & 81.0 & 71.9 & 61.8 & 86.5 & 72.9 & 67.9 & 15.9 & 26.2 & 8.1 \\ 
\textbf{Qwen3-0.6B} & 79.8 & 69.1 & 60.2 & 86.0 & 71.1 & 65.7 & 16.1 & 26.7 & 8.2 \\ \midrule
\end{tabular}}
\label{tab:llm}
\end{table*}}

\subsection{Analysis of robustness for pretrained models}

To assess the generalizability of KEC, we evaluate it under three different vision-language backbones. These backbones vary significantly in architecture and capacity. Specifically, CLIP ResNet-50 utilizes a convolutional network. CLIP ViT-B/16 introduces a Transformer-based model, and ImageBind-Huge is a recent multi-modal foundation model that jointly embeds multiple modalities, providing a stronger and more generalized representation space.
As shown in~\autoref{tab:backbone}, KEC consistently achieves the best performance across all backbones and metrics, significantly outperforming both the vanilla CLIP (K-Means) and TAC. These results validate the transferability of our hierarchical knowledge construction approach across a wide spectrum of model architectures and data domains.

\subsection{Knowledge construction with various LLMs}

We investigate the effect of using different LLMs in the knowledge construction stage of KEC. Specifically, we compare several state-of-the-art LLMs from diverse providers, including: Claude-3-7-Sonnet (Anthropic), Gemini-2.0-Flash (Google), Deepseek-V3, Qwen-Turbo, Qwen2.5-7B-Instruct, and Qwen-3-0.6B. The latter three represent models with relatively small parameters, allowing us to evaluate the feasibility of lightweight deployment. KEC is robust with open-source and small LLMs (even with a 0.6B model). As shown in~\autoref{tab:llm}, all models yield competitive performance, confirming the robustness of our hierarchical concept-attribute construction method.
These results confirm that our method does not rely on any specific LLM backbone and can generalize across different model families and deployment constraints.

\section{Constructed Knowledge Space}
\label{sec:know}

\subsection{Reducing Redundancy Compared to TAC.}

\begin{figure*}[t]
    \centering
    \includegraphics[width=\linewidth]{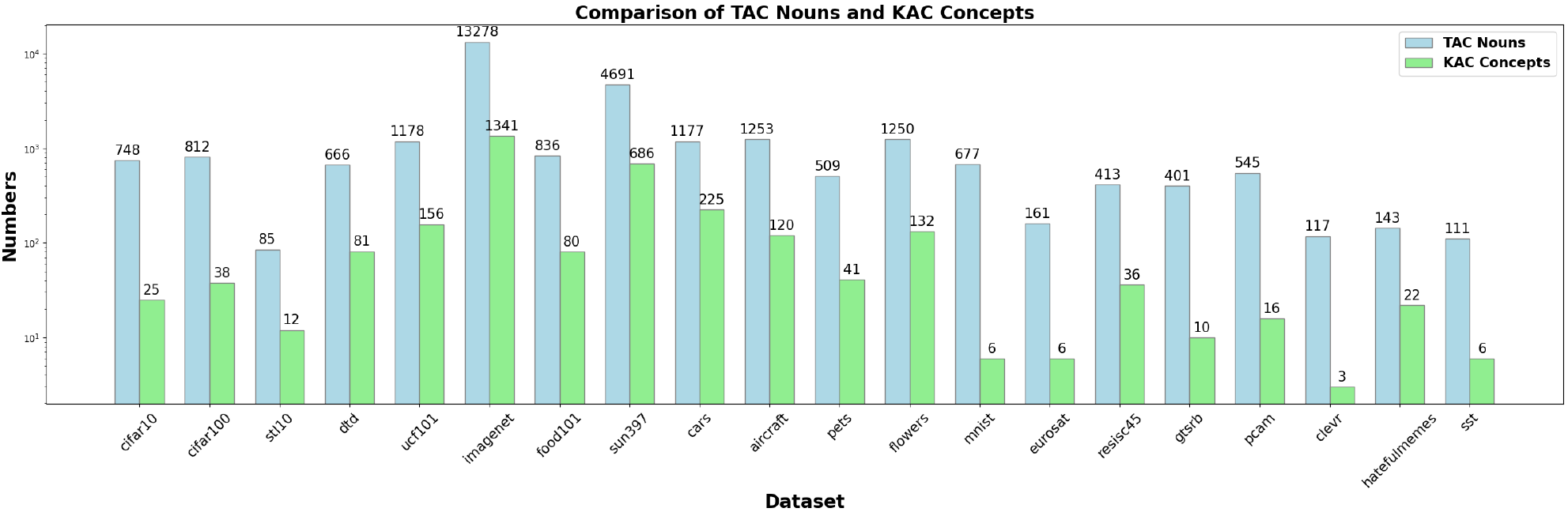}
    \caption{KEC effectively reduces semantic redundancy compared to naive use of textual knowledge.}
    \label{fig:number}
\end{figure*}

To better understand the effectiveness and efficiency of hierarchical knowledge construction, we compare the size of the knowledge space produced by TAC and our method (KEC) across 20 datasets. As shown in~\autoref{fig:number}, KEC significantly reduces the number of textual elements in the constructed knowledge, transforming hundreds or even thousands of nouns in TAC into a compact set of concepts. For example, on ImageNet, the number of textual tokens drops from 13,278 to 1,341, and similar reductions are observed across datasets such as SUN397 (from 4,691 to 686) and Aircraft (from 1,253 to 120). This demonstrates that our hierarchical knowledge construction mechanism effectively eliminates semantic redundancy and consolidates overlapping or overly fine-grained nouns into unified, higher-level concepts. Based on these representative concepts, KEC further explores discriminative attributes to assist in clustering.

\begin{figure*}[!h]
    \centering
    \includegraphics[width=6.7in]{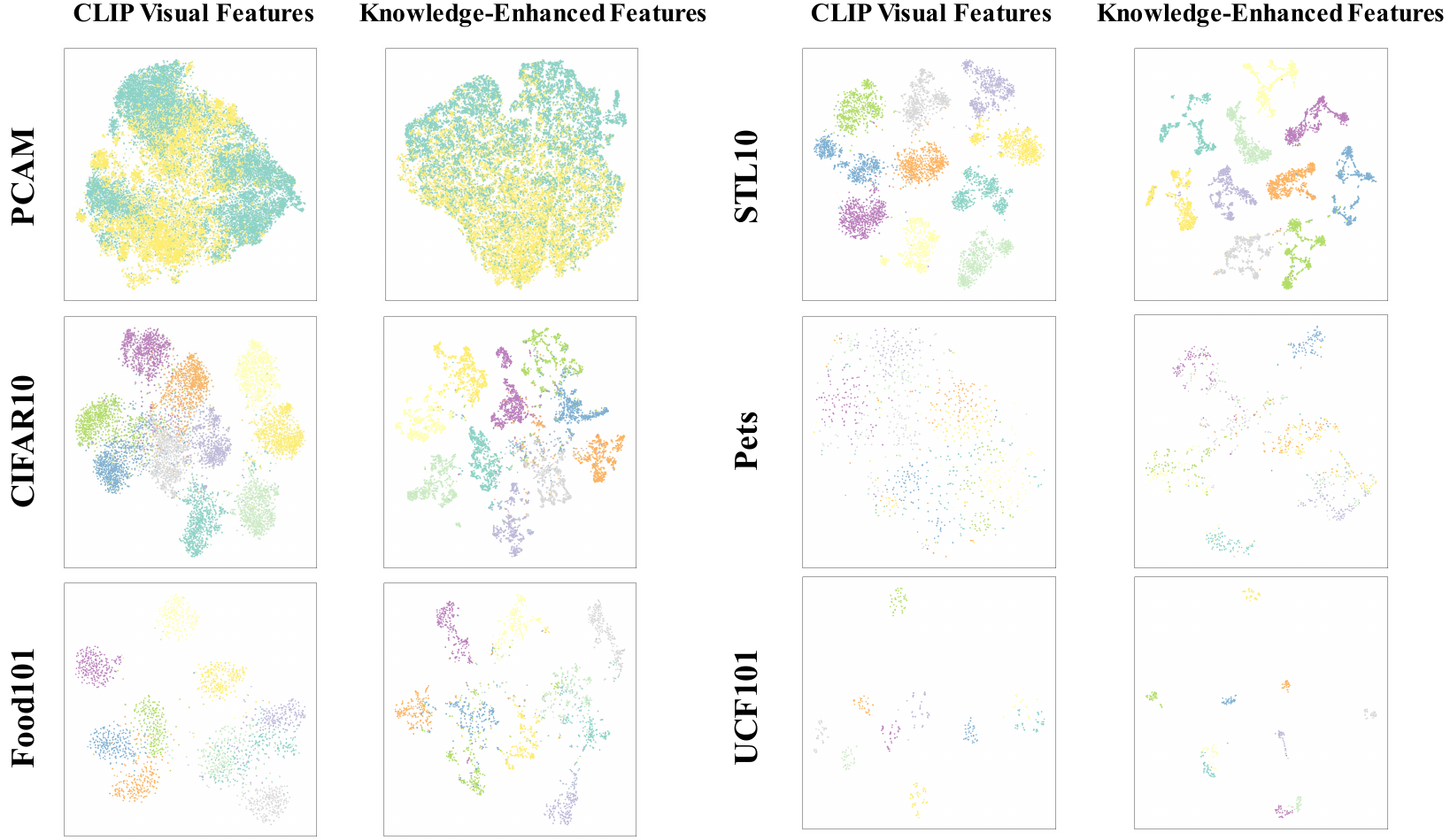}
    \caption{t-SNE visualization of features before and after knowledge enhancement across six representative datasets. The left column shows CLIP visual features, and the right column shows knowledge-enhanced features obtained by KEC. For datasets with large numbers of classes, only the first 10 classes are visualized for clarity.}
    \label{fig:tsne}
\end{figure*}

\subsection{Visualization of knowledge enhanced features.}

To qualitatively assess the effect of knowledge-enhanced features, we visualize the feature before and after enhancement using t-SNE across six datasets: PCAM, STL10, CIFAR10, Pets, Food101, and UCF101. 
The left panels show the raw CLIP visual features, while the right panels show the corresponding knowledge-enhanced features generated by KEC. For datasets with a large number of classes, we visualize only the first 10 categories to improve clarity.
As shown in~\autoref{fig:tsne}, the knowledge-enhanced features provided by KEC lead to better semantic separability, with tighter intra-cluster cohesion and clearer inter-cluster boundaries. This effect is especially prominent on fine-grained datasets like Pets and Food101, where textual knowledge helps refine class-specific distinctions that may not be evident from visual features alone.
These results provide intuitive evidence that KEC injects structured, semantically meaningful textual guidance into the visual space, thus aiding downstream clustering performance.

{\renewcommand{\arraystretch}{1.}
\begin{table*}[h]
\centering
\caption{Examples of the hierarchical knowledge constructed by KEC. We show 20 representative concepts, 10 Uni-Concept and Bi-Concept attributes, respectively. These knowledge elements are automatically derived via LLMs and serve to inject structured semantic guidance for clustering.}
\resizebox{\linewidth}{!}{
\begin{tabular}{p{0.15\linewidth}|p{0.85\linewidth}}
\hline
\textbf{Dataset} & \textbf{Hierarchical Knowledge} \\
\hline
CIFAR10 & \textbf{Concepts:} \\
 & Commercial and Utility Vehicles, Bird Species, Water and Land Transportation, aviation and air travel, Automobiles, Maritime Vessels, Musical Heritage, Equestrian Sports, Ecological Diversity of Amphibians and Reptiles, Pets, Feline and Animal, Antelope and Deer Species, emergency response vehicles, Equine Management and Culture, Aircraft, Dog Breeds, Maritime Vessels and Shipping, Fish Species, aviation and avian agility, Plant Ecology \\
 & \textbf{Uni-Concept Attributes:} \\
 & Prominent cargo area or flatbed design, Feathers covering the body, colors on animal fur or skin, Beak shape and size, Streamlined fuselage with a pointed nose, Large and swept-back wings with distinctive winglets, wheels or tracks on vehicles, Large hull structure with a streamlined shape, sails or masts or other rigging elements, Bright and high-visibility color schemes \\ 
 & \textbf{Bi-Concept Attributes:} \\
 & Feathered or fur-covered bodies, Color Variation, Mode of travel, Number of wheels, cargo or shipping containers on the vessels, Size and domestication level, representations and cultural symbols, Body structure and limb configuration, Color scheme and markings, landmark achievements \\
\hline
Flowers & \textbf{Concepts:} \\
 & Ornamental Plants and Flowering Species, Yellow Flora, Berry Cultivation, Flowering Plants, Primulaceae and Related Plants, Fritillaria, Flora of Tropical and Subtropical Regions, Wildflowers and Mosses, Wildflowers and Mosses, Irises and Associated Flora, Irises and Associated Flora, Goldenrod Species, Squash and Related Plants, Dianthus, Hymenoptera and Related Insects, Phyllostachys, Hibiscus and Mallow Plants, Native North American Flora, Penstemon and Antirrhinum Species, Arum Lilies and Egrets \\
 & \textbf{Uni-Concept Attributes:} \\
 & Vibrant and varied color palettes, Thick and serrated leaves, Waxy texture, a complex and organized structure, Black and white photographic style, Furry or feathery texture, palmate or lobed leaves with a more pronounced venation pattern, ovate or elliptical leaves with a smoother margin, Large white and petal-like ray florets, Colorful and diverse petal shapes and patterns \\ 
 & \textbf{Bi-Concept Attributes:} \\
 & Flower shape and color, Colorful and intricate floral patterns, Leaf shape and arrangement, Physical Form, Number of blooms, Color and texture, Vibrancy of Colors, Visual Style, abstract or stylized expressions, like paintings or sculptures \\
\hline
\end{tabular}}
\label{tab:examples}
\end{table*}}

\subsection{Examples of the hierarchical knowledge.}

To provide a clearer understanding of the knowledge constructed by our method, we present examples of hierarchical knowledge extracted from two representative datasets.
As shown in~\autoref{tab:examples}, the hierarchical knowledge includes three parts:
(1) \textbf{Concepts}, abstracted from semantically similar noun clusters; (2) \textbf{Uni-Concept Attributes}, which are representative and visually grounded features associated with individual concepts; and (3) \textbf{Bi-Concept Attributes}, which are contrastive properties mined between pairs of similar concepts to enhance fine-grained discrimination. 

For each dataset, we showcase 20 representative concepts and 10 attributes for each type. For example, in CIFAR10, concepts such as `Emergency Response Vehicles' or `Ecological Diversity of Amphibians and Reptiles' are abstracted from low-level nouns. Uni-concept Attributes like `Prominent cargo area' or `Beak shape and size' highlight core visual traits, while Bi-concept attributes such as `Mode of travel' or `Limb configuration' aid in separating fine-grained categories.

Similarly, in the Flowers dataset, concepts cover a wide range of botanical categories, while Uni-Concept attributes emphasize textural and morphological features (e.g., “Waxy texture” or “Serrated leaves”). Bi-Concept attributes, such as “Visual Style” or “Number of blooms,” further help differentiate visually similar floral species. These structured knowledge components form the basis of our knowledge-enhanced feature representation. 

This qualitative evidence underscores the semantic richness and interpretability of the constructed knowledge, which not only improves clustering performance but also contributes to the explainability of the clustering process.

\section{Limitation and Future Work}
\label{sec:limit}

While our proposed KEC achieves strong and consistent performance across diverse datasets and clustering settings, we highlight several limitations that also point toward promising future extensions:

\paragraph{Concept quantity may not always be larger than the target cluster number.}
In most cases, the number of concepts generated by our framework exceeds the target number of clusters, preventing overly coarse semantic summarization for concepts. However, in rare cases, such as CIFAR-100, only 38 concepts are formed. It is fewer than the dataset’s 100 fine-grained classes. Notably, this number still exceeds the 20 coarse-grained categories defined in CIFAR-20, which uses the same image set. This observation suggests that our method tends to generate semantically meaningful concepts, even without hard constraints. In future work, incorporating user-defined granularity or constraints into the concept construction process could support more adaptive and goal-driven clustering.

\paragraph{Limited improvement for high-level or subjective categories.}
The improvement is relatively modest in datasets where categories are defined by abstract or high-level semantics, such as HatefulMemes. The labels reflect whether an image is offensive. KEC may lack the precise guidance needed to construct appropriate textual knowledge. This points to an opportunity to actively involve users in knowledge construction, for example by supplying domain-specific vocabulary, mapping terms to visual regions, or refining the concept-attribute hierarchy. Such human-in-the-loop strategies could enable more accurate alignment between textual semantics and high-level visual concepts.

\paragraph{Potential for Interactive and Customized Knowledge Construction.}
KEC is modularity and flexibility. Allowing users to define or adjust certain concepts and attributes, or to specify domain-specific requirements, could further improve clustering performance. Exploring such interactive, customized knowledge construction mechanisms represents a compelling future direction.

In summary, we present a hierarchical knowledge construction method that opens up new possibilities for leveraging textual space in image clustering. While our current system operates automatically and performs well across tasks, future extensions incorporating user input, domain adaptation, and controllable knowledge granularity could further enhance its applicability and effectiveness.

\end{document}